\pdfoutput=1
\documentclass[11pt]{article}

\usepackage[letterpaper,margin=1in]{geometry}
\usepackage[round]{natbib}
\usepackage{graphicx}
\usepackage{amsmath}
\usepackage{amsfonts}
\usepackage{amssymb}
\usepackage{algorithm}
\usepackage{algorithmic}
\usepackage{booktabs}
\usepackage{pifont}
\usepackage{xcolor}
\usepackage{url}
\usepackage[hidelinks]{hyperref}

\newtheorem{theoryprop}{Proposition}

\newcommand{\cmark}{\ding{51}}

\hypersetup{
  pdftitle={Fed-BRDECS: Privacy-Preserving and Heterogeneity-Aware Federated Deep Embedded Clustering},
  pdfauthor={Haemin Park, Diego Klabjan, Martin W. Braun, Xiuqi Li, Balakrishnan Ananthanarayanan}
}

\title{Fed-BRDECS: Privacy-Preserving and Heterogeneity-Aware\\Federated Deep Embedded Clustering}
\author{
Haemin Park$^{1}$ \quad
Diego Klabjan$^{1}$ \quad
Martin W. Braun$^{2}$\\
Xiuqi Li$^{2}$ \quad
Balakrishnan Ananthanarayanan$^{2}$\\[0.5em]
\small $^{1}$Northwestern University \quad $^{2}$Intel Corporation\\
\small \texttt{haemin.park1@northwestern.edu} \quad \texttt{d-klabjan@northwestern.edu}\\
\small \texttt{martin.w.braun@intel.com} \quad \texttt{xiuqi.li@intel.com} \quad \texttt{a.bala.krishnan@intel.com}
}
\date{}

\begin{document}
\maketitle

\begin{abstract}
Federated deep clustering seeks to learn clustering-friendly representations from decentralized unlabeled data while preserving client privacy. However, Deep Embedded Clustering (DEC)-style objectives depend on global soft-assignment statistics that require clients to reveal their sensitive information. We propose Fed-BRDECS, a privacy-preserving and heterogeneity-aware federated deep embedded clustering framework. Fed-BRDECS replaces the globally normalized clustering objective with a locally computable sample-stability loss, avoiding the transmission of local soft-assignment distributions. To tackle non-IID client distributions, we introduce prediction-balanced sampling, which oversamples locally rare predicted clusters without requiring ground-truth labels, and centroid-level restarting, which periodically refreshes biased or inactive centroids. Experiments on image and text clustering benchmarks show that Fed-BRDECS consistently outperforms representative federated clustering and deep clustering baselines under both IID and non-IID partitions. We further demonstrate its applicability to federated time-series anomaly detection, where it improves reconstruction-based detectors without adding inference-time cost.
\end{abstract}

\section{INTRODUCTION}


Clustering is a fundamental unsupervised learning task, but in many applications data are distributed across institutions or devices and cannot be centrally collected due to privacy and ownership constraints \citep{jain2010data,rieke2020future, nguyen2021federated, pfeifer2024federated, presotto2022fedclar}. Federated learning (FL) provides a natural paradigm for this setting by enabling multiple clients to collaboratively train a model while keeping raw data local \citep{mcmahan2017communication}. Extending clustering to the federated setting is therefore important for privacy-preserving unsupervised representation learning, where labels are unavailable and data heterogeneity is often severe \citep{dennis2021heterogeneity, garst2024federated}. However, directly applying centralized deep clustering methods to FL is nontrivial.

DEC \citep{xie2016unsupervised} and its improved variant IDEC \citep{ijcai2017p243} are the dominant modern approaches to deep clustering, especially in the centralized regime. After autoencoder pretraining, they jointly refine latent representations and cluster assignments using soft assignments and an auxiliary target distribution. In FL, this target requires dataset-level assignment frequencies, and reproducing it requires either exchanging local cluster statistics, which exposes client composition, or constructing targets from incomplete local views, which can bias optimization under heterogeneity. 

Heterogeneity can further lead to biased centroid updates and inactive clusters \citep{dennis2021heterogeneity, pan2023machine}. For example, cluster-frequency-based objectives may over-emphasize locally dominant clusters and under-train rare clusters. Federated clustering can also suffer from cluster collapse and early saturation. Once the encoder and centroids become aligned with an imperfect early partition, subsequent training may reinforce the same assignments rather than correct them. Recent work identifies this phenomenon as a reclustering barrier \citep{miklautz2025breaking}. In FL, this early saturation can be further amplified because client updates are computed from partial, heterogeneous views and then averaged at the server.

We propose Fed-BRDECS (Federated Balanced-Sampling and Restarted Deep Embedded Clustering with Sample Stability), a federated deep clustering framework with three components. These enhancements address the privacy issues inherent in directly applying DEC to FL, while simultaneously tackling the core challenges of federated deep clustering. 

First, it replaces the globally normalized IDEC clustering objective with a locally computable sample-stability loss inspired by DECS \citep{cheng2024deep}. Since this objective is computed using only local samples and current centroids, clients do not need to transmit sensitive soft-assignment distributions. Second, to mitigate local imbalance under heterogeneous FL, it uses a prediction-based balanced sampler that oversamples each client's locally rare predicted clusters using the model's own assignments. This acts as an unsupervised form of class-balanced importance sampling. Third, periodic federated restarting recomputes centroids in the current embedding space to recover collapsed clusters.

DEC has several applications, with outlier detection being one of the most common. We use Fed-BRDECS as the building block for federated outlier detection and validate on time-series data from both public benchmarks and a proprietary sensor dataset. Since many detectors are built on deep embedded clustering structures \citep{zong2018deep, obied2023deep}, our framework can be easily applied to DEC-based or AE-based federated time-series anomaly detection algorithms. This demonstrates that the proposed framework is not limited to standard clustering benchmarks, but can also serve as a privacy-preserving representation-learning backbone for downstream unsupervised tasks.

Our contributions are summarized as follows.
\begin{enumerate}
    \item We propose \textbf{Fed-BRDECS}, a federated deep embedded clustering framework for heterogeneous unsupervised federated learning. Unlike prior federated or distributed clustering methods that exchange local clustering information, Fed-BRDECS enables IDEC-style joint representation and clustering refinement using a locally computable sample-stability loss without transmitting local assignment statistics.

    \item We introduce two mechanisms to improve federated deep clustering under heterogeneous client distributions: a \textbf{prediction-based balanced sampler} that mitigates local cluster imbalance using pseudo-labels, and an \textbf{iterative restarting strategy} that refreshes centroids to reduce early commitment and cluster collapse.

    \item We demonstrate consistent empirical improvements over representative federated clustering and deep clustering baselines, with ablation studies validating the contribution of stability loss, balanced sampling, and restarting.

    \item We evaluate Fed-BRDECS on \textbf{federated time-series anomaly detection} as a practical stress test for unsupervised federated representation learning, showing that the proposed framework can improve reconstruction-based detectors under heterogeneous client time-series distributions without adding inference-time cost.
\end{enumerate}

\section{RELATED WORK}

\paragraph{Deep clustering.}
DEC \citep{xie2016unsupervised} is a pioneering work showing that a representation can be learned directly from a clustering objective. IDEC \citep{ijcai2017p243} adds a reconstruction term to preserve local structure, and DCN \citep{yang2017towards} similarly connects representation learning with k-means style clustering. A common limitation of DEC-based methods is their reliance on manually constructed pseudo-targets, which may reinforce early assignment errors and depend on dataset-level assignment statistics. DECS \citep{cheng2024deep} replaces pseudo-targets with sample-stability optimization, which we adopt to eliminate global assignment statistics. BRB \citep{miklautz2025breaking} addresses early saturation through reclustering and soft parameter resets. Fed-BRDECS adapts these ideas to heterogeneous FL using local stability optimization and centroid-level restarting without parameter resets.

\begin{table}[t]
\centering
\caption{Comparison with representative federated clustering methods. A check mark indicates that the method explicitly supports the corresponding property. ``Privacy-preserving'' means that clients do not transmit local cluster assignments or soft-assignment matrices.}
\label{tab:fed_clustering_comparison}
\vspace{0.2cm}
\resizebox{0.65\columnwidth}{!}{%
\begin{tabular}{lccc}
\toprule
Method & Privacy-preserving & Deep clustering & Heterogeneity-aware \\
\midrule
k-FED & \cmark & & \\
DWF k-Means & & & \\
FKM & & & \\
FFCM & & & \\
F-DEC & & \cmark & \\
CCFC & \cmark & \cmark & \\
Fed-BRDECS & \cmark & \cmark & \cmark \\
\bottomrule
\end{tabular}%
}
\end{table}

\paragraph{Federated clustering.}
Federated clustering has mostly focused on centroid-style algorithms over fixed features rather than deep representation learning. k-FED \citep{dennis2021heterogeneity} is a one-shot method in which each client uploads its local cluster centers only once. FKM \citep{garst2024federated} extends this paradigm to multiple communication rounds. Other methods address specific limitations of federated k-means: DWF k-Means \citep{holzer2023dynamically} introduces dynamic weighting, Fork \citep{bravo2026robust} improves robustness, and FedDP k-Means \citep{scott2025differentially} incorporates differential privacy. In contrast to these hard-assignment approaches, FFCM \citep{stallmann2022towards} adapts fuzzy c-means to federated learning, allowing samples to have soft membership across multiple clusters. However, these methods still perform centroid-based clustering on raw or precomputed features rather than jointly learning deep representations, and they also exchange explicit local cluster statistics. More recently, efforts have emerged to apply deep clustering to FL. F-DEC \citep{9534268} directly applies DEC to FL by optimizing a KL-based clustering loss across clients, and FDEC \citep{XU2025113963} attempts to achieve robustness through hybrid federated averaging. CCFC \citep{liu2024ccfc} incorporates a cluster-contrastive loss to learn clustering-friendly representations. Still, these deep frameworks do not explicitly resolve core FL challenges such as centroid collapse, data heterogeneity, and privacy leakage tied to transmitting local soft assignment statistics. Table~\ref{tab:fed_clustering_comparison} summarizes these differences.

\paragraph{Imbalance in federated and unsupervised learning.}
Heterogeneity and imbalance are central challenges in FL. Supervised FL methods commonly address them through local data rebalancing, loss reweighting, or aggregation adjustment \citep{wang2021addressing, seol2023performance, chung2025multi, park2026classgrouped}. However, these methods rely on labels or label-distribution information, which are unavailable in unsupervised federated clustering. In unlabeled settings, related works---particularly in centralized environments---utilize inferred groups or pseudo-clusters to mitigate imbalance. For instance, some approaches balance unlabeled datasets using cluster IDs as pseudo-classes, while others partition majority classes into subclusters for imbalanced classification \citep{suayama2025reducing, lim2024data}. We adapt this concept to federated clustering. However, rather than treating predicted cluster assignments as supervised targets, we use them strictly to define a local sampling distribution paired with an oversampling strategy.

\section{METHOD}

\subsection{Preliminaries: DEC/IDEC}
\label{sec:preliminaries}

DEC and IDEC combine representation learning with centroid-based clustering to map raw data into a semantically meaningful latent space. Let $\mathcal{D}=\{x_i\}_{i=1}^{N}$ be a centralized dataset. An encoder $f_\theta:\mathcal{X}\to\mathcal{Z}$ maps an input $x_i$ to a latent representation $z_i=f_\theta(x_i) \in \mathcal{Z}$ and a decoder $g_\phi:\mathcal{Z}\to\mathcal{X}$ reconstructs the input, encouraging the latent representation to preserve local structures. The framework maintains $C$ trainable centroids $\{\mu_j\}_{j=1}^{C}\subset\mathcal{Z}$. Standard IDEC operates in three stages:

\begin{enumerate}
    \item \textbf{Autoencoder Pretraining:} The network minimizes reconstruction loss to establish a meaningful initial embedding space:
    \begin{equation}\label{eq:train_autoencoder}
        \min_{\phi, \theta} \mathcal{L}_{\mathrm{rec}} = \frac{1}{N}\sum_{i=1}^{N} \left\Vert x_i-g_\phi(f_\theta(x_i))\right\Vert_2^2.
    \end{equation}

    \item \textbf{Centroid Initialization:} Centroid-based clustering (usually, k-Means++) is applied to the pretrained embeddings $\{z_i=f_\theta(x_i)\}_{i=1}^{N}$ to initialize the centroids $\{\mu_j\}_{j=1}^{C}$ where $\phi,\theta$ are learned through~\eqref{eq:train_autoencoder}.

    \item \textbf{Joint Clustering Refinement:} The encoder, decoder, and centroids are jointly optimized. For each sample, a soft assignment $q_{ij}$ is computed using a Student-$t$ kernel:
    \begin{equation}\label{eq:student_t}
        q_{ij} = \frac{\left(1+\Vert z_i-\mu_j\Vert_2^2/\alpha\right)^{-\frac{\alpha+1}{2}}}{\sum_{\ell=1}^{C} \left(1+\Vert z_i-\mu_\ell\Vert_2^2/\alpha\right)^{-\frac{\alpha+1}{2}}},
    \end{equation}
    where $\alpha>0$ denotes the degrees-of-freedom parameter. A large value of $q_{ij}$ indicates that $z_i$ is closer to centroid $\mu_j$ relative to the other centroids. An auxiliary target distribution $P$ is then constructed to sharpen confident assignments after constructing soft cluster frequencies $f_j = \sum_{i=1}^{N}q_{ij}$:
    \begin{equation}\label{eq:target}
        p_{ij} = \frac{q_{ij}^2/f_j}{\sum_{\ell=1}^{C} q_{i\ell}^2/f_\ell}.
    \end{equation}
    The parameters are updated via stochastic gradient descent by minimizing the joint loss:
    \begin{equation}
        \mathcal{L}_{\mathrm{IDEC}} = \mathcal{L}_{\mathrm{rec}} + \lambda_{\mathrm{clu}} \operatorname{KL}(P\Vert Q),
    \end{equation}
    where $\lambda_{\mathrm{clu}}$ balances reconstruction and cluster refinement. DEC only uses clustering loss without reconstruction loss for clustering refinement.
\end{enumerate}

\subsection{Problem Formulation: Federated Deep Embedded Clustering}
\label{sec:problem_formulation}

We consider an unsupervised federated learning setting with $K$ clients. Client $k$ owns an unlabeled private dataset $\mathcal{D}_k = \{x_{k,i}\}_{i=1}^{n_k}$. The global dataset is defined as $\mathcal{D} = \bigcup_{k=1}^{K} \mathcal{D}_k$ and the total number of samples is $N = \sum_{k=1}^{K} n_k$. The goal is to learn a single global clustering model that partitions the distributed dataset into $C$ clusters.

\paragraph{Privacy issue in FL.}
Directly applying IDEC to FL introduces a severe problem. Computing the global target distribution requires aggregating dataset-wide soft frequencies $f_j$. In FL, this forces each client $k$ to compute and upload local soft counts $f_{k,j} = \sum_{x_i\in\mathcal{D}_k}q_{ij}$ to the server. This mechanism leaks sensitive information regarding local cluster compositions and relies on stale global statistics between periodic server refreshes. These limitations motivate Fed-BRDECS to replace the global KL target with a locally computable sample-stability objective.

\paragraph{Problem formulation.}
Following the deep embedded clustering framework, the global model is denoted by $\Theta = \{\theta, \phi, \mu_1, \dots, \mu_C\}$. It consists of an encoder $f_\theta : \mathcal{X} \to \mathcal{Z}$, a decoder $g_\phi : \mathcal{Z} \to \mathcal{X}$, and $C$ centroids $\{\mu_j\}_{j=1}^{C} \subset \mathcal{Z}$ in the latent space $\mathcal{Z}$. Given a trained model, the cluster assignment of a sample $x$ is determined by the nearest centroid in the latent space, $\hat{y}(x)=\arg\min_{j \in \{1,\dots,C\}} \| f_\theta(x) - \mu_j \|_2^2$. Since training is unsupervised, these assignments are model-induced pseudo-cluster labels rather than ground-truth labels. Because these local assignments can still leak privacy-sensitive information, we focus on a federated clustering design that entirely eliminates the transmission of local assignment statistics.

The federated objective can be written as the weighted average of local unsupervised objectives:
\[
\min_{\Theta} \;
\mathcal{F}(\Theta)
=
\sum_{k=1}^{K}
r_k
\mathcal{F}_k(\Theta; \mathcal{D}_k),
\]
where $\mathcal{F}_k$ is the local clustering objective evaluated on client $k$'s data, and $r_k=\frac{n_k}{N}$.

This formulation highlights the central difficulty of federated deep embedded clustering. The model must learn both the representation and the centroids from decentralized, unlabeled, and potentially non-IID data, while avoiding reliance on centralized assignment statistics. Fed-BRDECS addresses this difficulty by replacing the globally normalized IDEC target objective with a local sample-stability loss, correcting local prediction imbalance through a prediction-based balanced sampler, and periodically reinitializing centroids through federated restarting.

\subsection{Algorithm: Fed-BRDECS}
\label{sec:algorithm}

\begin{algorithm}[!t]
\caption{Fed-BRDECS}
\label{alg:fed-brdecs}
\raggedright
\begin{algorithmic}[1]
\REQUIRE Client datasets $\{\mathcal D_k\}_{k=1}^{K}$, number of clusters $C$, pretraining rounds $T_{\mathrm{pre}}$, restart iterations $R$, refinement rounds $T$, local steps $E$, sampler period $T_{\mathrm{samp}}$, loss weight $\lambda_{\mathrm{stab}}$, stability parameter $\lambda_{\mathrm{var}}$
\ENSURE Global model $\Theta=\{\theta,\phi,\mu_1,\ldots,\mu_C\}$
\item[] \textbf{(Step 1) Federated representation pretraining.}
\STATE Server initializes the global model $\Theta^0$.
\FOR{$t=1,\ldots,T_{\mathrm{pre}}$}
    \STATE Server broadcasts $\Theta^{t-1}$ to clients.
    \FOR{$k=1,\ldots,K$}
        \STATE Initialize local model $\Theta_k^{t,0}\leftarrow \Theta^{t-1}$.
        \STATE Perform $E$ local SGD steps on $\mathcal L_{\mathrm{rec}}^{(k)}$.
        \STATE Upload $\Theta_k^{t,E}$ to the server.
    \ENDFOR
    \STATE Server aggregates: $\Theta^{t} = \sum_{k} r_k \Theta_k^{t,E}$.
\ENDFOR
\item[] \textbf{(Step 2) Federated centroid initialization.}
\STATE $\{\mu_j\}_{j=1}^{C} \leftarrow \mathrm{k\text{-}FED}\big(\{f_{\theta^{T_{\mathrm{pre}}}}(\mathcal D_k)\}_{k=1}^{K}\big)$; update $\Theta^{T_\mathrm{pre}}$ with these centroids.
\item[] \textbf{(Step 3) Restarted clustering refinement.}
\STATE Initialize $\Theta^{0,T} \leftarrow \Theta^{T_\mathrm{pre}}$.
\FOR{$r=1,\ldots,R$}
    \STATE Initialize $\Theta^{r,0} \leftarrow \Theta^{r-1, T}$.
    \STATE Recompute $\{\mu_j^r\}_{j=1}^{C} \leftarrow \mathrm{k\text{-}FED}\big(\{f_{\theta^{r,0}}(\mathcal D_k)\}_{k=1}^{K}\big)$.
    \STATE Set $\Theta^{r,0} \leftarrow \{\theta^{r,0},\phi^{r,0},\mu_1^{r},\ldots,\mu_C^{r}\}$.
    \FOR{$t=1,\ldots,T$}
        \STATE Server broadcasts $\Theta^{r,t-1}$ to clients.
        \FOR{$k=1,\ldots,K$}
            \STATE Initialize local model $\Theta_{k,0}^{r, t}\leftarrow \Theta^{r,t-1}$.
            \IF{$t \equiv 0 \pmod{T_{\mathrm{samp}}}$}
                \STATE Compute sampler weights $w_{k,i}$.
            \ENDIF
            \STATE Perform $E$ local SGD steps on $\mathcal L_k = \mathcal L_{\mathrm{rec}}^{(k)} + \lambda_{\mathrm{stab}} \mathcal L_{\mathrm{stab}}^{(k)}$ using prediction-balanced mini-batches.
            \STATE Upload $\Theta_{k,E}^{r,t}$ to the server.
        \ENDFOR
        \STATE Server aggregates: $\Theta^{r,t}= \sum_{k} r_k\Theta_{k,E}^{r,t}$.
    \ENDFOR
\ENDFOR
\RETURN $\Theta^{R,T}$
\end{algorithmic}
\end{algorithm}

Fed-BRDECS optimizes a federated deep clustering objective in three stages: federated representation pretraining, federated centroid initialization, and restarted clustering refinement. The first two stages follow the standard deep embedded clustering pipeline, while the refinement stage replaces IDEC's globally normalized target objective with local sample stability, prediction-balanced sampling, and periodic federated restarting. Algorithm~\ref{alg:fed-brdecs} summarizes the full procedure.

\paragraph{Federated representation pretraining.}
Each client minimizes the reconstruction loss in~\eqref{eq:train_autoencoder}
on its local data, and the server aggregates the updates by FedAvg for
$T_{\mathrm{pre}}$ rounds.


\paragraph{Federated centroid initialization.}
After pretraining, the server initializes the latent centroids using a federated clustering routine on the client embeddings:
\[
\{\mu_j\}_{j=1}^{C}
\leftarrow
\mathrm{k\text{-}FED}
\left(
\{f_\theta(\mathcal{D}_k)\}_{k=1}^{K}
\right).
\]
The FL clustering algorithm k-FED \citep{dennis2021heterogeneity} plays the role of centralized k-Means initialization in DEC/IDEC, but it is performed without collecting raw data or local assignment information at the server. Each client computes local centroids from its embeddings and sends only these centroids to the server, which constructs the initial global centroids.

\paragraph{Restarted clustering refinement.}
The refinement stage is organized into $R$ restart iterations. Unlike BRB-style reclustering, Fed-BRDECS does not apply a soft reset to the model parameters. Instead, each restart performs a centroid-level federated reclustering step while keeping the representation network continuous across restarts. This design focuses on correcting centroid degeneration, inactive clusters, and heterogeneity-induced assignment imbalance.

At the beginning of restart $r$, the server simply recomputes the centroids by running k-FED on the current client embeddings, $\{\mu_j\}_{j=1}^{C} \leftarrow \mathrm{k\text{-}FED}(\{f_\theta(\mathcal D_k)\}_{k=1}^{K})$, warm-started from the previous centroids. This federated reclustering step relocates centroids to occupied regions of the current embedding space and can revive centroids that have become inactive under non-IID local updates.

Within each restart iteration, Fed-BRDECS performs $T$ rounds of federated refinement. On client $k$, soft assignments are computed from the local model using the Student-$t$ kernel with parameter $\alpha$:
\[
q_{kij}
=
\frac{
\left(
1+\|z_{k,i}-\mu_{k,j}\|_2^2/\alpha
\right)^{-\frac{\alpha+1}{2}}
}{
\sum_{\ell=1}^{C}
\left(
1+\|z_{k,i}-\mu_{k,\ell}\|_2^2/\alpha
\right)^{-\frac{\alpha+1}{2}}
},
\]
where $z_{k,i}=f_{\theta_k}(x_{k,i})$.
Unlike IDEC, Fed-BRDECS does not construct a target distribution from globally normalized cluster frequencies. Instead, each client evaluates sample stability, inspired by \citet{cheng2024deep}, from its local assignment probabilities.

Let $\tau_k\in(0,1)$ be the threshold selected by Otsu's method \citep{4310076} on the assignment probabilities observed at client $k$. We define the determinacy score
$d_{kij}=((q_{kij}-\tau_k)/\tau_k)^2$ if $q_{kij}<\tau_k$, and
$d_{kij}=((q_{kij}-\tau_k)/(1-\tau_k))^2$ otherwise.
For sample $x_{k,i}$, we define the average determinacy as $\bar d_{k,i}=\frac{1}{C} \sum_{j=1}^{C} d_{kij}$. Then the stability score is
\[
s_{k,i}
=
\bar d_{k,i}
-
\frac{\lambda_{\mathrm{var}}}{C}
\sum_{j=1}^{C}
\left(
d_{kij}-\bar d_{k,i}
\right)^2,
\]
where $\lambda_{\mathrm{var}}\ge0$ is a proportionality coefficient. The first term rewards assignment determinacy with respect to all centroids, while the second term regularizes the dispersion of determinacy values across centroids. The local stability loss over a mini-batch $\mathcal B_k$ is
\[
\mathcal L_{\mathrm{stab}}^{(k)}
=
1
-
\frac{1}{|\mathcal B_k|}
\sum_{x_{k,i}\in\mathcal B_k}
s_{k,i}.
\]
The local refinement objective is
\[
\mathcal L_k
=\mathcal L_{\mathrm{rec}}^{(k)}+
\lambda_{\mathrm{stab}}
\mathcal L_{\mathrm{stab}}^{(k)},
\]
where $\lambda_{\mathrm{stab}}$ controls the contribution of the stability loss. The reconstruction term is also retained during refinement, as in IDEC, to preserve local structure in the embedding space. The encoder, decoder, and centroids are updated jointly by local SGD between restart steps.

\paragraph{Prediction-balanced sampler.}
To address local prediction imbalance, each client periodically updates a prediction-based sampling distribution. Every $T_{\mathrm{samp}}$ rounds, client $k$ computes local soft cluster frequencies $f_{k,j}=\sum_{i=1}^{n_k} q_{kij}$ ($j=1,\dots,C$) and hard predicted assignments $\hat y_{k,i} = \arg\max_{j\in\{1,\dots,C\}} q_{kij}$.
The inverse predicted-cluster frequency is $\tilde w_{k,j} = 1/(f_{k,j}+\epsilon)$, and each sample receives weight
\[
w_{k,i}
=
\frac{
\tilde w_{k,\hat y_{k,i}}
}{
\frac{1}{C}
\sum_{\ell=1}^{C}
\tilde w_{k,\ell}
}.
\]
Mini-batches are then sampled with replacement according to probabilities proportional to $w_{k,i}$. Thus, samples assigned to locally rare predicted clusters are sampled more often, while samples assigned to locally dominant predicted clusters are downsampled. The predicted assignments are used only to define the sampling distribution and are not treated as supervised labels.

At refinement round $t$ of restart iteration $r$, the server broadcasts the current global model $\Theta^{r,t-1}$ to the clients. Each client performs $E$ local steps using the current balanced sampler and uploads its updated parameters $\Theta_{k,E}^{r,t}$. The server aggregates the returned models.

{
\subsection{Theoretical Analysis}
\label{sec:theory}

A privacy-aware alternative to the globally normalized DEC target is to let each client reconstruct the target using only its local soft cluster frequencies. Under heterogeneous client cluster proportions, however, this local reconstruction is not equivalent to the centralized DEC target and can even reverse the relative preference between two clusters. Appendix~\ref{appx:local_target_distortion} formalizes this effect and bounds both the target distortion and the resulting clustering-gradient bias. Thus, replacing the global DEC target with a locally computable stability objective avoids this source of heterogeneity-induced target mismatch.

The local stability objective avoids this dependence on global frequencies $f_j$ in \eqref{eq:target}, but it contains no explicit cluster-frequency balancing term. Consequently, centroids may become inactive, as formalized in
Appendix~\ref{appx:stability_collapse}. The following result explains why
such centroids are difficult to recover by gradient refinement and how
restarting addresses this failure mode.



\setcounter{theoryprop}{2}
\begin{theoryprop}[Inactive-centroid recovery]
\label{prop:dead_centroid}
Consider an inactive centroid $j$. If $q_{kij}\le\delta$ throughout $H$ local SGD steps with step size $\eta$, its displacement after FedAvg is at most
\[
\|\mu_j^{\mathrm{after}}-\mu_j^{\mathrm{before}}\|_2
\le
C_{\mathrm{stab}}\,\eta H\delta,
\]
where $C_{\mathrm{stab}}$ is independent of $\delta$. In contrast, suppose the restart step relocates each centroid into the convex hull of the current embeddings $Z$. Then, for every client, sample, and centroid,
\[
q_{kij}\ge
\frac1C\left(1+\frac{\operatorname{diam}(Z)^2}{\alpha}\right)^{-(\alpha+1)/2}>0
\]
Appendix~\ref{appx:dead_centroid_proof} gives all the details and assumptions.
\end{theoryprop}

Thus, the gradient signal of an inactive centroid vanishes with its assignment
mass, whereas restarting restores a strictly positive assignment mass by
relocating the centroid to the current embedding region.
}

\begin{figure*}[t]
    \centering
    \includegraphics[width=.95\textwidth]{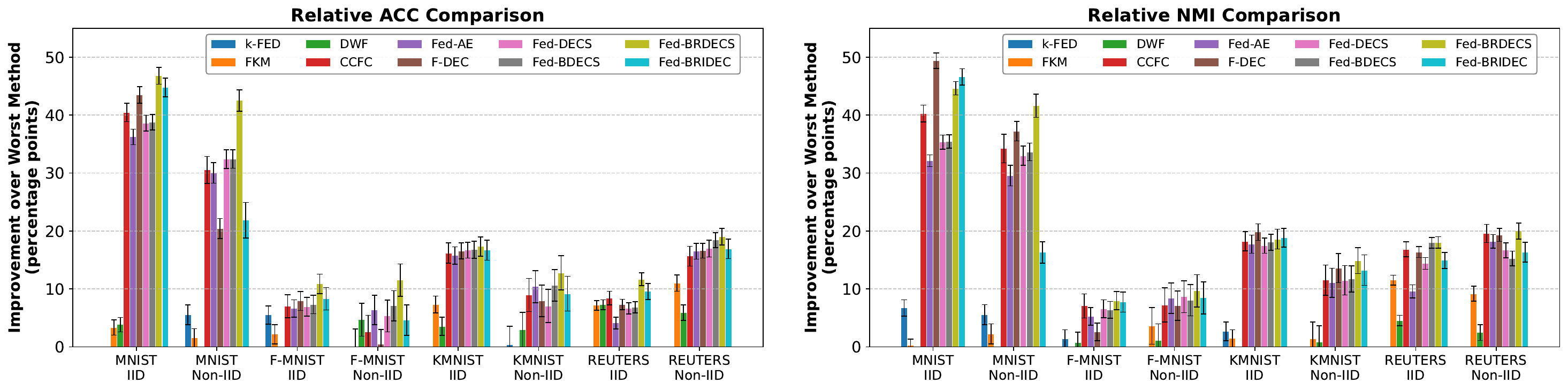}
    \caption{Performance gains relative to the worst-performing method for each dataset and partition. Bars show percentage-point improvements in ACC (left) and NMI (right), and error bars indicate one standard deviation over five runs. Higher is better. Lower performance of DEC-based refinement compared with Fed-AE indicates that the refinement stage can degrade the pretrained representation, especially under non-IID client distributions.}
    \label{fig:performance_comparison}
\end{figure*}

\section{EXPERIMENTS}

\subsection{Setup for Clustering on Public Datasets}
\label{sec:setup}

\paragraph{Datasets.}
We evaluate Fed-BRDECS on four standard deep clustering benchmarks: MNIST \citep{6296535}, Fashion-MNIST \citep{xiao2017fashion}, KMNIST \citep{clanuwat2018deep}, and REUTERS-10K \citep{lewis2004rcv1}. The three image datasets each contain ten ground-truth categories, while REUTERS-10K is a four-category text clustering benchmark commonly used in DEC/IDEC evaluations. We follow the standard preprocessing protocol used in deep embedded clustering methods \citep{xie2016unsupervised, ijcai2017p243, cheng2024deep}. Ground-truth labels are used only for federated distribution of samples to clients (partitioning) and final evaluation, and are never used during unsupervised training.

\paragraph{Federated partitioning.}
All experiments are conducted with $K=10$ clients. For each dataset, we evaluate two federation settings. In the IID setting, samples are uniformly distributed across clients. In the non-IID setting, client data distributions are generated using a Dirichlet partition with concentration parameter $\alpha=0.5$, as described in \citet{hsu2019measuring}. The non-IID setting tests whether each method can learn meaningful global clusters when clients observe different local cluster proportions.

\paragraph{Baselines.}
We compare Fed-BRDECS with three groups of baselines. The first group consists of federated centroid clustering methods over fixed features: k-FED \citep{dennis2021heterogeneity}, FKM \citep{garst2024federated}, and DWF k-Means \citep{holzer2023dynamically}. These methods do not learn deep representations.

The second group consists of deep federated clustering baselines. Fed-AE first trains an autoencoder in a federated manner and then applies k-FED to the learned embeddings, isolating the effect of representation pretraining without joint clustering refinement. F-DEC \citep{9534268} is a direct federated implementation of IDEC, where clients optimize IDEC reconstruction and the KL clustering objective under FedAvg. This baseline uses the same autoencoder backbone but retains the DEC/IDEC target distribution objective and, therefore, requires exchange of local clustering information to construct the global target distribution. CCFC \citep{liu2024ccfc} is a recent contrastive federated deep clustering method.

The third group consists of our ablation variants. Fed-DECS uses only the local sample-stability loss, corresponding to the stability component (denoted as $S$). Fed-BDECS adds prediction-balanced sampling ($S{+}B$). Fed-BRDECS is the full method with sample stability, prediction-balanced sampling, and centroid-level restarting ($S{+}B{+}R$). We also report Fed-BRIDEC, which applies balanced sampling and restarting to the IDEC-style objective, to separate the effect of our stability objective from the effect of the balancing and restarting mechanisms.

All deep clustering methods use the standard DEC autoencoder architecture
\citep{xie2016unsupervised,ijcai2017p243} and the same federated pretraining
protocol. Additional hyperparameters and training details are provided in
Appendix~\ref{appx:experiment_details}.

\paragraph{Evaluation metric.}
All clustering methods are evaluated by clustering accuracy (ACC) and Normalized Mutual Information (NMI), which are widely used in unsupervised clustering. Unless otherwise stated, we report the average over five independent runs with different random seeds.

\begin{figure}[t]
    \centering
    \includegraphics[width=0.9\columnwidth]{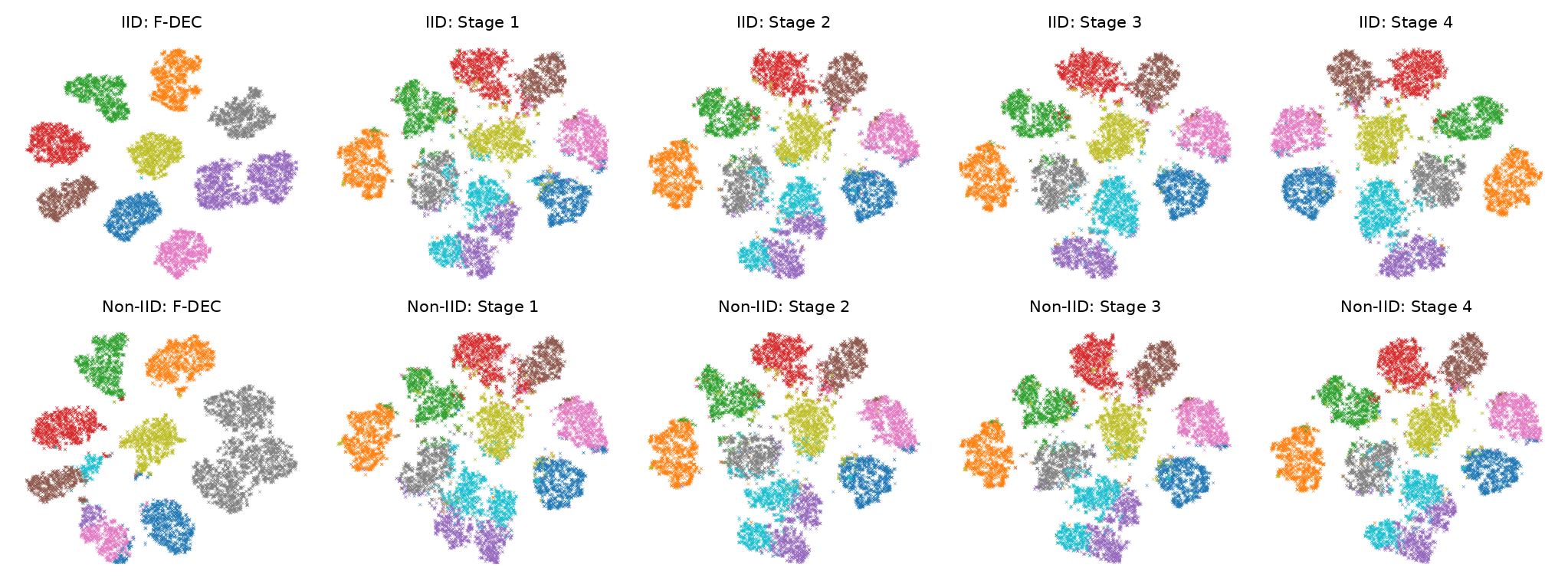}
    \caption{t-SNE visualization of learned MNIST embeddings under IID and non-IID federated partitions. The first column shows F-DEC, while the remaining columns show Fed-BRDECS across restart/refinement stages. Under non-IID partition, F-DEC exhibits clustering collapse, with two classes mixed into poorly separated regions. Fed-BRDECS mitigates this collapse through centroid-level restarting, progressively producing more compact and separated cluster structures.}
    \label{fig:mnist_tsne}
\end{figure}

\subsection{Clustering Results}
\label{sec:clustering_results}

In Figure~\ref{fig:performance_comparison}, results are shown as percentage-point improvements over the worst-performing method for each dataset, partition, and metric. Fed-BRDECS attains the highest ACC in all eight settings. The corresponding absolute ACC and NMI values are reported in Appendix~\ref{app:full_clustering_results}. Overall, Fed-BRDECS achieves the strongest and most consistent clustering performance across datasets and partition settings. Its advantage is particularly evident under non-IID distributions, indicating improved robustness to heterogeneous client data.

\paragraph{Effect of representation learning.}
The fixed-feature federated k-means methods are consistently weaker than deep clustering methods on most benchmarks (Figure~\ref{fig:performance_comparison}). For example, on MNIST-IID, the best fixed-feature baseline obtains $0.4914$, whereas Fed-AE reaches $0.8155$. This confirms that learning a clustering-friendly representation remains crucial in federated clustering. However, Fed-AE is still weaker than Fed-BRDECS, showing that autoencoder pretraining alone is insufficient without clustering-aware refinement. In several settings, particularly under non-IID partitions, F-DEC also underperforms Fed-AE, indicating that DEC-style refinement can degrade the pretrained representation.

\paragraph{Effectiveness of proposed components.}
F-DEC is a direct federated implementation of the IDEC objective. Fed-BRDECS improves over F-DEC by 6.47 percentage points in average ACC across the eight settings, with especially large gains under non-IID partitions. Further, the ablation ladder Fed-DECS $\rightarrow$ Fed-BDECS $\rightarrow$ Fed-BRDECS shows the contribution of each component. Fed-DECS, which uses only the stability loss, already performs competitively with F-DEC while avoiding the transmission of local cluster-frequency statistics. Adding prediction-balanced sampling and centroid-level restarting further improves ACC, indicating that our B+R components help correct biased or inactive centroids during federated refinement.

\paragraph{Robustness to heterogeneity.}
Fed-BRDECS is particularly effective under non-IID partitions. Averaged over the four non-IID settings, Fed-BRDECS obtains $0.6889$ ACC, outperforming other algorithms. This suggests that the combination of prediction-balanced sampling and centroid restarting improves robustness to heterogeneous local cluster proportions.

\paragraph{Embedding visualization.}
Figure~\ref{fig:mnist_tsne} visualizes the learned MNIST embeddings using t-SNE. Compared with F-DEC, Fed-BRDECS mitigates cluster collapse and cluster mix-up. The improvement is especially visible under the non-IID partition, where the baseline exhibits stronger cluster mixing, whereas Fed-BRDECS progressively separates clusters through prediction-balanced refinement and centroid-level restarting.

\paragraph{Ablation on the clustering-loss weight.}
Figure~\ref{fig:lambda_clu_sensitivity} in Appendix~\ref{app:implementation_details} examines the effect of $\lambda_{\mathrm{stab}}$ on MNIST and Fashion-MNIST. Under non-IID partitions, both datasets perform best at $\lambda_{\mathrm{stab}}=0.5$, whereas the IID results remain strong around $\lambda_{\mathrm{stab}}=1.0$; we use these values in the main experiments.

\subsection{Application to Federated Time-Series Anomaly Detection}
\label{sec:tsad}

Beyond standard clustering benchmarks, we evaluate Fed-BRDECS in federated time-series anomaly detection (TSAD), which provides a practical stress test for unsupervised federated representation learning. Many TSAD methods are reconstruction based. A model is trained to reconstruct normal time series, and anomalous time series are identified by a large reconstruction error \citep{malhotra2016lstm, audibert2020usad, tuli2022tranad}. In federated TSAD, different clients may correspond to different machines or operating environments, so normal patterns can be highly heterogeneous. As a result, a single reconstruction model trained by FedAvg may underrepresent rare normal regimes and produce false alarms.

\paragraph{Applying Fed-BRDECS to reconstruction-based TSAD.}
Fed-BRDECS can be applied to reconstruction-based TSAD backbones as a training-time module. If the detector already includes a latent clustering component, we directly replace its clustering objective with our sample-stability loss and add prediction-balanced sampling and centroid restarting. If the detector is purely reconstruction-based, we attach a temporary projection head and centroids to the latent representation, optimize the sample-stability objective jointly with reconstruction, and use the resulting pseudo-clusters for balanced sampling.

We instantiate this mechanism on LSTM-AE \citep{garg2021evaluation} and TranAD \citep{tuli2022tranad} for regularly sampled public benchmarks, attaching the clustering head to the encoder final hidden state and to the latent representation used by the reconstruction module, respectively. The proprietary sensor data are irregularly sampled multivariate time series, so we use CoFormer-AE \citep{wei2023compatible}, which replaces the LSTM encoder with CoFormer; TranAD and other architecture-specific TSAD variants are not directly compatible with this backbone. At inference time, the clustering head is discarded and outlier scores are computed using the original reconstruction error, so Fed-BRDECS adds no inference-time cost.

\paragraph{Experimental setup.}
We evaluate on two public multivariate TSAD benchmarks, SMD \citep{su2019robust} and PSM \citep{abdulaal2021practical}, following the federated evaluation setting of FedTADBench \citep{liu2022fedtadbench}, and on a proprietary irregularly sampled sensor dataset collected from multiple industrial machines. Each backbone is trained with FedAvg either with or without Fed-BRDECS. We report AUC-ROC and F1. Further dataset and implementation details are provided in Appendix~\ref{appx:tsad_impl}.

\begin{table}[t]
\centering
\caption{Federated time-series anomaly detection results. ``Baseline'' denotes the backbone trained with FedAvg, while ``+BRDECS'' additionally uses Fed-BRDECS during training.}
\label{tab:tsad_results}
\vspace{0.2cm}
\resizebox{0.65\columnwidth}{!}{%
\begin{tabular}{llcccc}
\toprule
& & \multicolumn{2}{c}{Baseline} & \multicolumn{2}{c}{+BRDECS} \\
\cmidrule(lr){3-4}
\cmidrule(lr){5-6}
Dataset & Backbone & AUC-ROC & F1 & AUC-ROC & F1 \\
\midrule
SMD    & LSTM-AE     & 0.5915 & 0.1413 & \textbf{0.6038} & \textbf{0.1422} \\
SMD    & TranAD      & 0.6335 & 0.1829 & \textbf{0.6396} & \textbf{0.1848} \\
PSM    & LSTM-AE     & 0.6063 & 0.4415 & \textbf{0.6121} & \textbf{0.4433} \\
PSM    & TranAD      & 0.5792 & \textbf{0.4451} & \textbf{0.5863} & 0.4425 \\
Sensor & CoFormer-AE & 0.5258 & 0.4503 & \textbf{0.5544} & \textbf{0.4629} \\
\bottomrule
\end{tabular}%
}
\end{table}

\paragraph{Results.}
Table~\ref{tab:tsad_results} shows that adding Fed-BRDECS improves AUC-ROC in all five evaluated dataset--backbone settings and improves F1 in four of five settings. These gains are modest but consistent, which is expected because Fed-BRDECS is used only as a training-time representation module, while the clustering components and balanced sampler may help mitigate heterogeneity in FL.

\section{CONCLUSION}

We presented Fed-BRDECS, a federated deep embedded clustering framework that replaces the globally normalized IDEC target with a locally computable sample-stability loss, so that clients do not transmit soft-assignment or cluster-frequency statistics during refinement, and that counters local imbalance and centroid collapse through prediction-balanced sampling and centroid-level restarting. Fed-BRDECS attains the highest ACC in all eight federated clustering settings, and it also serves as a training-time module for downstream unsupervised monitoring, improving reconstruction-based anomaly detectors without changing their inference procedure. It does not provide formal differential privacy, since model updates and k-FED centroid means are still communicated (Appendix~\ref{appx:privacy_scope}); adding formal privacy mechanisms for these centroids is a natural next step.

\subsection*{AI Use Statement}

In this work, we used generative AI tools for editing and sanitizing sentences to improve clarity, grammar, and readability.
We have not used generative AI tools for conceptual design, conducting experiments, data analysis, or generating new scientific claims, and automated code or dataset generation are not applicable to this work.
We have reviewed all AI-assisted work. Every revised sentence was manually verified by the authors to ensure that the original technical meaning, nuances, and factual accuracy were strictly preserved without introducing hallucinations or alterations to the findings.
We take responsibility for the final content of this work, including text, claims, or artifacts produced with the aid of generative AI.

\bibliography{fedbrdecs}


\clearpage
\appendix

{
\section{THEORETICAL ANALYSIS: PROOFS AND ADDITIONAL DETAILS}
\label{appx:theory_proofs}

To simplify the Student-$t$ soft assignment equation \eqref{eq:student_t}, define the unnormalized Student-$t$ affinity
\[
 u_{ij} :=  \left(1+\frac{\|z_i-\mu_j\|_2^2}{\alpha}\right)^{-\frac{\alpha+1}{2}}.
\]
Then \eqref{eq:student_t} can be written as
\[
 q_{ij}=\frac{u_{ij}}{\sum_{\ell=1}^{C}u_{i\ell}}.
\]
We also use
\[
 h_{ij}
 :=
 \frac{\alpha+1}{\alpha}
 \frac{z_i-\mu_j}{1+\|z_i-\mu_j\|_2^2/\alpha}.
\]
To simplify \eqref{eq:target}, let $a=(a_1,\ldots,a_C)$ be any positive frequency vector and define
\[
 p_{ij}(a)
 :=
 \frac{q_{ij}^2/a_j}{\sum_{\ell=1}^{C}q_{i\ell}^2/a_\ell},
 \qquad
 p_i(a):=\bigl(p_{i1}(a),\ldots,p_{iC}(a)\bigr).
\]
Then \eqref{eq:target} becomes $p_{ij}(f)$ when $f=(f_1,\ldots,f_C)$ is the global soft-frequency vector. For client $k$, we write $f^{(k)}=(f_{k,1},\ldots,f_{k,C})$ for its local soft-frequency vector. For probability vectors $p,p'$, we use $\operatorname{TV}(p,p')=\frac12\|p-p'\|_1$.

\subsection{Auxiliary lemmas}

\paragraph{Lemma A.1 (Student-$t$ derivatives).}
For any sample $i$ and clusters $j,\ell$, the Student-$t$ kernel satisfies
\[
 \nabla_{z_i}\log u_{ij}=-h_{ij},
 \qquad
 \nabla_{\mu_j}\log u_{ij}=h_{ij},
\]
\[
 \nabla_{z_i}q_{i\ell}
 =
 q_{i\ell}
 \left(-h_{i\ell}+\sum_m q_{im}h_{im}\right),
\]
\[
 \nabla_{\mu_j}q_{i\ell}
 =
 q_{i\ell}\bigl(\mathbf 1[\ell=j]-q_{ij}\bigr)h_{ij}.
\]
In addition,
\[
 \|h_{ij}\|_2\le c_\alpha:=\frac{\alpha+1}{2\sqrt\alpha}.
\]

\paragraph{Proof.}
From the definition of $u_{ij}$,
\[
 \log u_{ij}
 =
 -\frac{\alpha+1}{2}
 \log\left(1+\frac{\|z_i-\mu_j\|_2^2}{\alpha}\right).
\]
Differentiating with respect to $z_i$ and $\mu_j$ gives the first two identities. The derivatives of $q_{i\ell}$ follow from differentiating $\log q_{i\ell}=\log u_{i\ell}-\log\sum_m u_{im}$. Finally, with $r=\|z_i-\mu_j\|_2$,
\[
 \|h_{ij}\|_2
 =
 \frac{\alpha+1}{\alpha}
 \frac{r}{1+r^2/\alpha},
\]
and $r/(1+r^2/\alpha)$ is maximized at $r=\sqrt\alpha$, where it equals $\sqrt\alpha/2$. \hfill$\square$

\paragraph{Lemma A.2 (Multiplicative reweighting).}
Let $p=(p_1,\ldots,p_C)$ be a probability vector with $p_j>0$ for all $j$. Let $\omega_j>0$ and define
\[
 p_j'=\frac{p_j\omega_j}{\sum_\ell p_\ell\omega_\ell}.
\]
If
\[
 \Delta=\log\frac{\max_j\omega_j}{\min_j\omega_j},
\]
then
\[
 \operatorname{TV}(p',p)
 \le
 \tanh\left(\frac{\Delta}{4}\right).
\]
For $C=2$, the bound can be attained.

\paragraph{Proof.}
Set $a=\min_j\omega_j$ and $b=\max_j\omega_j$. Let $J$ have distribution $p$, define $W=\omega_J$, and let $m=\mathbb E[W]$. By the definition of $p'$, we have
\[
 \operatorname{TV}(p',p)
 =
 \frac{\mathbb E|W-m|}{2m}.
\]
Since $W\in[a,b]$,
\[
 \mathbb E|W-m|
 \le
 \frac{2(m-a)(b-m)}{b-a}.
\]
It follows that
\[
 \operatorname{TV}(p',p)
 \le
 \frac{(m-a)(b-m)}{m(b-a)}.
\]
The last expression is maximized at $m=\sqrt{ab}$. Substitution gives
\[
 \operatorname{TV}(p',p)
 \le
 \frac{\sqrt{b/a}-1}{\sqrt{b/a}+1}
 =
 \tanh\left(\frac14\log\frac ba\right).
 \qquad\square
\]

\subsection{Local-target distortion under local DEC-target reconstruction}
\label{appx:local_target_distortion}

For any positive frequency vector $a=(a_1,\ldots,a_C)$, define the DEC target probability obtained using $a$ by
\[
p_{ij}(a)
:=
\frac{q_{ij}^2/a_j}{\sum_{\ell=1}^{C}q_{i\ell}^2/a_\ell},
\qquad j=1,\ldots,C,
\]
and define the complete target vector for sample $i$ as
\[
p_i(a):=\bigl(p_{i1}(a),\ldots,p_{iC}(a)\bigr)\in\Delta_C.
\]
In particular, $f=(f_1,\ldots,f_C)$ denotes the global soft-frequency vector, while $f^{(k)}=(f_{k,1},\ldots,f_{k,C})$ denotes client $k$'s local soft-frequency vector. Thus, $p_i(f)$ is the centralized DEC target for sample $i$, whereas $p_i(f^{(k)})$ is the target reconstructed using only client $k$'s local frequencies.

\setcounter{theoryprop}{0}
\begin{theoryprop}[Local-target distortion]
\label{prop:local_target}
Let $\widehat\pi_{k,j}=f_{k,j}/n_k$, $\widehat\pi_j=f_j/N$, $\omega_{k,j}=\widehat\pi_j/\widehat\pi_{k,j}$, and
\[
\Delta_k=\log\frac{\max_j\omega_{k,j}}{\min_j\omega_{k,j}}.
\]
For every sample $i$ of client $k$,
\[
p_{ij}\!\left(f^{(k)}\right)
=
\frac{p_{ij}(f)\omega_{k,j}}
{\sum_\ell p_{i\ell}(f)\omega_{k,\ell}},
\]
and consequently
\[
\operatorname{TV}\!\left(p_i(f^{(k)}),p_i(f)\right)
\le
\tanh\!\left(\frac{\Delta_k}{4}\right).
\]
Moreover,
\[
\left\|
\nabla_{z_i}\operatorname{KL}\!\left(p_i(f^{(k)})\Vert q_i\right)
-
\nabla_{z_i}\operatorname{KL}\!\left(p_i(f)\Vert q_i\right)
\right\|_2
\le
2c_\alpha\tanh\!\left(\frac{\Delta_k}{4}\right),
\]
where $c_\alpha=(\alpha+1)/(2\sqrt\alpha)$. The locally reconstructed target is identical to the global target, $p_i(f^{(k)})=p_i(f)$, if and only if the normalized local and global soft-frequency vectors are identical, $\widehat\pi_k=\widehat\pi$. Equivalently, this holds if and only if $\Delta_k=0$. If $\Delta_k>0$, the mismatch can change both the target probabilities and the clustering gradient.
\end{theoryprop}

\paragraph{Proof.}
Because $f_j=N\widehat\pi_j$ and $f_{k,j}=n_k\widehat\pi_{k,j}$,
\[
 \frac{q_{ij}^2}{f_{k,j}}
 =
 \frac{N}{n_k}
 \frac{\widehat\pi_j}{\widehat\pi_{k,j}}
 \frac{q_{ij}^2}{f_j}.
\]
The common factor $N/n_k$ cancels when the target is normalized, giving
\[
 p_{ij}(f^{(k)})
 =
 \frac{p_{ij}(f)\omega_{k,j}}
 {\sum_\ell p_{i\ell}(f)\omega_{k,\ell}}.
\]
Applying Lemma A.2 with $p=p_i(f)$ and $\omega=\omega_k$ yields
\[
 \operatorname{TV}\!\left(p_i(f^{(k)}),p_i(f)\right)
 \le
 \tanh\left(\frac{\Delta_k}{4}\right).
\]

When differentiating the DEC objective, the auxiliary target is held fixed. By Lemma A.1,
\[
 \nabla_{z_i}\operatorname{KL}(p_i\Vert q_i)
 =
 \sum_j(p_{ij}-q_{ij})h_{ij}.
\]
Let $p'=p_i(f^{(k)})$ and $p=p_i(f)$. Then
\begin{align*}
 &\left\|\nabla_{z_i}\operatorname{KL}(p'\Vert q_i)
 -\nabla_{z_i}\operatorname{KL}(p\Vert q_i)\right\|_2\\
 &\qquad\le
 \sum_j|p_j'-p_j|\,\|h_{ij}\|_2
 \le
 c_\alpha\|p'-p\|_1\\
 &\qquad=
 2c_\alpha\operatorname{TV}(p',p)
 \le
 2c_\alpha\tanh\left(\frac{\Delta_k}{4}\right).
\end{align*}
The centroid-gradient bound follows in the same way. The equality $p_i(f^{(k)})=p_i(f)$ holds exactly when $\omega_{k,j}$ is constant over $j$. Since both $\widehat\pi_k$ and $\widehat\pi$ sum to one, this is equivalent to $\widehat\pi_k=\widehat\pi$. Also,
\[
 \frac{p_{ij}(f^{(k)})}{p_{i\ell}(f^{(k)})}
 =
 \frac{p_{ij}(f)}{p_{i\ell}(f)}
 \frac{\omega_{k,j}}{\omega_{k,\ell}}.
\]
Therefore, if $p_{ij}(f)>p_{i\ell}(f)$, the local target reverses this ordering whenever
\[
 1<\frac{p_{ij}(f)}{p_{i\ell}(f)}<\frac{\omega_{k,\ell}}{\omega_{k,j}}.
\]
\hfill$\square$

Proposition~\ref{prop:local_target} shows that replacing the global DEC frequencies with client-specific frequencies does not recover the same target loss function. It produces a different target obtained by reweighting cluster $j$ by
\[
\omega_{k,j}=\frac{\widehat\pi_j}{\widehat\pi_{k,j}}.
\]
Therefore, a cluster that is rare on client $k$ is upweighted compared with the global DEC target, while a cluster that is common on that client is downweighted. The quantity $\Delta_k$ measures the spread of these reweighting factors. 

The preference-reversal condition $(1<\frac{p_{ij}(f)}{p_{i\ell}(f)}<\frac{\omega_{k,\ell}}{\omega_{k,j}})$ further shows that this effect is not limited to a change in gradient magnitude. Local reconstruction can change which centroid receives the larger target probability for the same sample. Thus, under heterogeneity, constructing the DEC target independently from local frequencies can make different clients optimize systematically different clustering directions. This motivates Fed-BRDECS to remove the frequency-normalized target from local refinement rather than approximate the global DEC target using local frequencies.

\subsection{Balance-agnostic stability objective}
\label{appx:stability_collapse}

The stability score $s_{k,i}$ for the $i$-th sample on client $k$ is given by
\[
s_{k,i}
=
\bar d_{k,i}
-
\frac{\lambda_{\mathrm{var}}}{C}
\sum_{j=1}^{C}(d_{kij}-\bar d_{k,i})^2,
\]
where $\lambda_{\mathrm{var}}\ge0$ is the variance-regularization coefficient. It controls how strongly dispersion among the centroid-wise determinacy values $\{d_{kij}\}_{j=1}^{C}$ is penalized.

Unlike the DEC target objective, the stability objective does not explicitly depend on cluster frequencies. It therefore encourages confident sample-level assignments without directly enforcing balanced use of the centroids. As a result, a training can achieve a small stability loss even when most or all samples are assigned to the same centroid. The following proposition formalizes this limitation and shows that the stability objective alone does not rule out inactive centroids.

\begin{theoryprop}[Balance-agnostic stability]
\label{prop:stability_collapse}
For fixed $\tau_k\in(0,1)$,
\[
0\le \mathcal L_{\mathrm{stab}}^{(k)}\le 1+\frac{\lambda_{\mathrm{var}}}{4}.
\]
Moreover, for any fixed encoder,
\[
\inf_{\{\mu_j\}}\mathcal L_{\mathrm{stab}}^{(k)}=0
\]
can be achieved by centroids in which all samples are assigned to a single centroid while the remaining $C-1$ centroids become inactive.
\end{theoryprop}

\paragraph{Proof.}

For $q<\tau$, the normalized deviation $(q-\tau)/\tau$ lies in $[-1,0)$, whereas for $q\ge\tau$, $(q-\tau)/(1-\tau)$ lies in $[0,1]$. Hence $d_{kij}\in[0,1]$ for every $i$ and $j$. Therefore, $\bar d_{k,i}\in[0,1]$, and the variance of $\{d_{kij}\}_{j=1}^{C}$ is bounded above by $1/4$. It follows that
\[
-\frac{\lambda_{\mathrm{var}}}{4}
\le
s_{k,i}
\le
1,
\]
which implies
\[
0
\le
\mathcal L_{\mathrm{stab}}^{(k)}
\le
1+\frac{\lambda_{\mathrm{var}}}{4}.
\]

For the second part, fix the encoder and $\mu_1$. For each $j\ge2$, set $\mu_j=\mu_1+m v_j$, where $v_j$ is a unit vector. Since the dataset is finite, letting $m\to\infty$ gives
\[
q_{i1}\to1,
\qquad
q_{ij}\to0
\quad (j\ge2)
\]
uniformly over the samples. Consequently, $d_{kij}\to1$ for all $j$, and therefore $s_{k,i}\to1$ and $\mathcal L_{\mathrm{stab}}^{(k)}\to0$. Thus, the infimum can be approached by configurations in which all samples are assigned to a single centroid while the remaining centroids become inactive. 
\hfill$\square$

\subsection{Proof and strengthened form of Proposition~\ref{prop:dead_centroid}}
\label{appx:dead_centroid_proof}

\paragraph{Assumptions for Proposition~\ref{prop:dead_centroid}.}
During the analyzed local-update window, each threshold $\tau_k$ is held fixed and satisfies $\tau_k\in[\tau_{\min},1-\tau_{\min}]$ for some $\tau_{\min}>0$. Local centroid updates use plain SGD with step size $\eta$. For the restart statement, each refreshed centroid is assumed to lie in the convex hull of the current embeddings. This condition is satisfied by Lloyd updates when empty clusters are reseeded at input embeddings.

\paragraph{Proof.}
We first bound the gradient with respect to an inactive centroid. Let $\phi(q_i)$ be continuously differentiable and suppose
\[
\left|
\frac{\partial \phi}{\partial q_{i\ell}}
\right|
\le G
\]
for every $\ell$. By Lemma~A.1,
\[
\nabla_{\mu_j}q_{i\ell}
=
q_{i\ell}
\bigl(\mathbf 1[\ell=j]-q_{ij}\bigr)h_{ij}.
\]
Therefore,
\begin{align*}
\nabla_{\mu_j}\phi(q_i)
&=
\sum_{\ell}
\frac{\partial\phi}{\partial q_{i\ell}}
q_{i\ell}
\bigl(\mathbf 1[\ell=j]-q_{ij}\bigr)h_{ij}\\
&=
q_{ij}
\sum_{\ell\ne j}
q_{i\ell}
\left(
\frac{\partial\phi}{\partial q_{ij}}
-
\frac{\partial\phi}{\partial q_{i\ell}}
\right)
h_{ij}.
\end{align*}
Using $\|h_{ij}\|_2\le c_\alpha$ gives
\[
\|\nabla_{\mu_j}\phi(q_i)\|_2
\le
2c_\alpha G q_{ij}(1-q_{ij})
\le
2c_\alpha G q_{ij}.
\]

We now apply this bound to the stability score. Since $d_{kij}\in[0,1]$ and $\tau_k\in[\tau_{\min},1-\tau_{\min}]$,
\[
\left|
\frac{\partial d_{kij}}
{\partial q_{kij}}
\right|
\le
\frac{2}{\tau_{\min}}.
\]
Moreover,
\[
\frac{\partial s_{k,i}}
{\partial d_{kij}}
=
\frac{1}{C}
-
\frac{2\lambda_{\mathrm{var}}}{C}
\bigl(d_{kij}-\bar d_{k,i}\bigr),
\]
and hence
\[
\left|
\frac{\partial s_{k,i}}
{\partial d_{kij}}
\right|
\le
\frac{1+2\lambda_{\mathrm{var}}}{C}.
\]
Thus,
\[
\left|
\frac{\partial s_{k,i}}
{\partial q_{kij}}
\right|
\le
G,
\qquad
G:=
\frac{2(1+2\lambda_{\mathrm{var}})}
{C\tau_{\min}}.
\]

The reconstruction loss does not depend on the centroids. Therefore, for a mini-batch $\mathcal B_{k,h}$ at local step $h$, we have

\begin{equation}\label{eq:bound_loss_batch_1}
\left\|
\nabla_{\mu_j}
\mathcal L_k(\mathcal B_{k,h})
\right\|_2
\le
\frac{2c_\alpha G\lambda_{\mathrm{stab}}}
{|\mathcal B_{k,h}|}
\sum_{i\in\mathcal B_{k,h}}
q_{kij}^{(h)}.
\end{equation}

Under the assumption $q_{kij}^{(h)}\le\delta$, for every client, sample, and local iterate in the window, \eqref{eq:bound_loss_batch_1} becomes
\[
\left\|
\nabla_{\mu_j}
\mathcal L_k(\mathcal B_{k,h})
\right\|_2
\le
2c_\alpha G\lambda_{\mathrm{stab}}\delta.
\]
A plain-SGD step with step size $\eta$ therefore moves the local centroid by at most $2c_\alpha G\lambda_{\mathrm{stab}}\eta\delta$. Summing over $H$ local steps gives
\[
\|\mu_{k,j}^{H}-\mu_j^{0}\|_2
\le
2c_\alpha G\lambda_{\mathrm{stab}}\eta H\delta.
\]
Since FedAvg forms a convex combination of the client centroids, the same bound holds for the aggregated centroid. Hence
\[
\|\mu_j^{\mathrm{after}}-\mu_j^{\mathrm{before}}\|_2
\le
2c_\alpha G\lambda_{\mathrm{stab}}\eta H\delta,
\]
so Proposition~\ref{prop:dead_centroid} holds with
\[
C_{\mathrm{stab}}
=
2c_\alpha G\lambda_{\mathrm{stab}}.
\]

For the restart statement, suppose each refreshed centroid belongs to $\operatorname{conv}(Z)$, where $Z$ is the current embedding set. Since $z_i\in Z$ and the diameter of a convex hull is equal to the diameter of its generating set,
\[
\|z_i-\mu_j\|_2
\le
D_Z,
\qquad
D_Z:=\operatorname{diam}(Z).
\]
It follows that
\[
u_{ij}
\ge
\left(
1+\frac{D_Z^2}{\alpha}
\right)^{-\frac{\alpha+1}{2}}.
\]
Also, $u_{i\ell}\le1$ for every $\ell$, so
\[
\sum_{\ell=1}^{C}u_{i\ell}\le C.
\]
Therefore,
\[
q_{ij}
=
\frac{u_{ij}}
{\sum_{\ell=1}^{C}u_{i\ell}}
\ge
\frac{1}{C}
\left(
1+\frac{D_Z^2}{\alpha}
\right)^{-\frac{\alpha+1}{2}}
=:\kappa>0.
\]
This proves the post-restart lower bound.
\hfill$\square$

\paragraph{A stronger cumulative-mass bound.}
The same argument gives a stronger form that does not require a common upper bound $\delta$ at every local step. Define the average assignment mass received by centroid $j$ on client $k$ at step $h$ as
\[
\bar q_{k,j}^{(h)}
=
\frac{1}{|\mathcal B_{k,h}|}
\sum_{i\in\mathcal B_{k,h}}
q_{kij}^{(h)},
\]
and let
\[
\delta_{j,h}
=
\max_k\bar q_{k,j}^{(h)}.
\]
Applying the same gradient bound at each step and summing over the local trajectory yields
\[
\|\mu_j^{H}-\mu_j^{0}\|_2
\le
2c_\alpha G\lambda_{\mathrm{stab}}\eta
\sum_{h=0}^{H-1}\delta_{j,h}.
\]
Thus the total movement of centroid $j$ is controlled by the cumulative assignment mass it receives during local optimization. The bound in Proposition~\ref{prop:dead_centroid} follows immediately when $\delta_{j,h}\le\delta$ for every $h$, since
\[
\sum_{h=0}^{H-1}\delta_{j,h}
\le
H\delta.
\]
}

\section{ADDITIONAL METHOD DETAILS}
\label{appx:method_details}

\subsection{k-FED Centroid Initialization and Restarting}
\label{appx:kfed_restart}

Fed-BRDECS uses k-FED \citep{dennis2021heterogeneity} for both initial centroid construction and periodic centroid restarting. We use k-FED because it constructs global centroids from uploaded local cluster means without requiring clients to transmit local soft assignment statistics or cluster-count statistics, which may reveal sensitive information about client-specific data composition.

After federated autoencoder pretraining, each client embeds its local data using the current encoder and solves a local clustering subproblem in the latent space. Following k-FED, the client sends only the resulting local cluster means to the server without sending cluster assignment statistics. The server then applies the k-FED server-side aggregation procedure to the collection of uploaded local means to obtain the global centroids. This procedure replaces the centralized k-means initialization used in DEC/IDEC with a one-shot federated centroid construction step.

The same procedure is used during restarting. At the beginning of each restart iteration, the server does not reset or perturb the encoder or decoder parameters as in BRB \citep{miklautz2025breaking}. Instead, clients recompute local cluster means from their current embeddings, and the server reruns the k-FED aggregation step to refresh the global centroids. The model parameters are kept unchanged across restarts, while the centroids are relocated according to the current federated embedding geometry. This design allows Fed-BRDECS to correct biased or inactive centroids induced by heterogeneous client updates without disrupting the learned representation.

Formally, at restart $r$, clients compute embeddings using the current encoder $f_{\theta^{r,0}}$ and the server updates
\[
\{\mu_j^r\}_{j=1}^{C} \leftarrow \mathrm{k\text{-}FED} \left( \{f_{\theta^{r,0}}(\mathcal D_k)\}_{k=1}^{K} \right).
\]
The model parameters are then set to
\[
\Theta^{r,0} = \{\theta^{r,0},\phi^{r,0},\mu_1^r,\ldots,\mu_C^r\},
\]
and federated refinement proceeds from this centroid-refreshed model.

\subsection{Privacy Scope of Fed-BRDECS}
\label{appx:privacy_scope}

Fed-BRDECS is privacy-preserving in the specific sense of reducing local assignment-statistic disclosure. In DEC/IDEC-style federated refinement, local soft-assignment matrices, soft-count vectors, or cluster-frequency statistics directly encode client-specific cluster composition. Such composition information can be sensitive in practical applications, for example when local clusters correspond to defect types, disease subpopulations, or machine operating regimes.

Fed-BRDECS does not provide formal differential privacy. Like other prototype-based federated clustering methods, it still communicates model updates and local centroid means during k-FED initialization and restarting. However, during clustering refinement, Fed-BRDECS avoids repeated transmission of local soft assignments, soft-count vectors, and predicted-cluster histograms. Existing work \citep{scott2025differentially} applies differential privacy to mitigate this leakage, but DP-noised cluster counts may still reveal client-level composition. Appendix~\ref{appx:dp_counts} further discusses why adding DP noise to cluster counts may fail to hide local composition unless the noise magnitude is large enough to substantially degrade the utility of the statistics.

\subsection{Why DP-Noised Cluster Counts Still Reveal Local Composition}
\label{appx:dp_counts}

A natural alternative to avoiding local assignment statistics is to transmit differentially private noisy cluster counts. We argue that this only partially addresses the privacy issue considered in this work. Differential privacy primarily protects individual-record participation, whereas our concern is client-level cluster-composition disclosure. In federated clustering, a local cluster-count vector directly encodes the client's data composition, which can be sensitive in applications such as industrial defect monitoring, where defect-type proportions may reveal process quality, machine conditions, or yield-related information.

Consider client $k$ with $n_k$ samples and $C$ clusters. Let
\[
c_k=(c_{k,1},\ldots,c_{k,C})\in\mathbb{N}^{C},
\qquad
\sum_{j=1}^{C}c_{k,j}=n_k,
\]
denote its local hard cluster-count vector, and let
\[
\pi_k=\frac{c_k}{n_k}
\]
be the corresponding local cluster-composition vector. For DEC/IDEC-style federated refinement, the analogous soft-count vector is
\[
f_k=(f_{k,1},\ldots,f_{k,C}),
\qquad
f_{k,j}=\sum_{i=1}^{n_k}q_{kij},
\]
whose normalized version similarly reveals the client's soft cluster composition.

Suppose a client applies the Laplace mechanism to its count vector:
\[
\widetilde c_k=c_k+\eta,
\qquad
\eta_j\sim \mathrm{Laplace}\left(\frac{\Delta}{\varepsilon}\right),
\]
where $\Delta$ is the $\ell_1$-sensitivity of the histogram query. The server can estimate the client's local composition by normalizing the noisy counts:
\[
\widetilde \pi_k=\frac{\widetilde c_k}{n_k}.
\]
Then
\[
\mathbb{E}\|\widetilde\pi_k-\pi_k\|_1
=
\frac{1}{n_k}\mathbb{E}\|\eta\|_1
=
\frac{C\Delta}{n_k\varepsilon}.
\]
Therefore, for moderate or large local datasets, the normalized noise can be very small unless $\varepsilon$ is chosen extremely small. For example, when $C=10$, $n_k=5{,}000$, $\Delta=1$, and $\varepsilon=1$, the expected $\ell_1$ error in the recovered composition is only
\[
\frac{10}{5000}=0.002.
\]
Thus, even after adding standard DP noise to the count vector, the server can still recover a highly accurate approximation of the client's local cluster composition.

This issue can also be understood through group privacy. Differential privacy protects neighboring datasets that differ in one record, but two clients with different cluster compositions may differ in many records. Let $\pi$ and $\pi'$ be two possible local composition vectors. Transforming a histogram with composition $\pi$ into one with composition $\pi'$ requires
\[
m=\frac{n_k}{2}\|\pi-\pi'\|_1
\]
sample replacements. By group privacy, an $\varepsilon$-DP mechanism provides only $m\varepsilon$-level indistinguishability for this composition-level change. Hence, to hide two compositions separated by $\delta=\|\pi-\pi'\|_1$, we need
\[
\frac{n_k\varepsilon\delta}{2}=O(1),
\qquad\text{or equivalently}\qquad
\varepsilon=O\left(\frac{1}{n_k\delta}\right).
\]
For a fixed composition gap $\delta$, this requires $\varepsilon=O(1/n_k)$. Under the Laplace mechanism, the corresponding noise scale becomes
\[
\frac{\Delta}{\varepsilon}=O(n_k),
\]
meaning that the noise added to each count is on the same order as the client's dataset size. Such noise would substantially corrupt the count vector and degrade its utility for federated k-means updates or DEC/IDEC target distribution construction.

Therefore, DP-noised cluster counts exhibit a sharp privacy--utility tradeoff for the composition-disclosure threat considered in this work. Moderate noise preserves clustering utility but still reveals an accurate estimate of local cluster composition; noise large enough to obscure that composition makes the statistics unreliable for clustering. Fed-BRDECS avoids this disclosure channel by not transmitting local soft-assignment matrices, local cluster-frequency vectors, or predicted-cluster histograms during refinement.

\section{EXPERIMENT DETAILS}
\label{appx:experiment_details}

\subsection{Datasets and Models}
\label{app:datasets_models}

\paragraph{Image datasets.}
For MNIST \citep{6296535}, Fashion-MNIST \citep{xiao2017fashion}, and KMNIST \citep{clanuwat2018deep}, we use the official test split of each dataset. Each image is flattened into a vector before being fed into the autoencoder. The encoder follows the standard DEC-style multilayer perceptron architecture
\[
784 \rightarrow 500 \rightarrow 500 \rightarrow 2000 \rightarrow 10,
\]
where the final layer gives a 10-dimensional latent representation. The decoder is symmetric. The number of clusters is set to the number of ground-truth classes, i.e., $C=10$. Labels are used only for partitioning and evaluation, not during training.

\paragraph{REUTERS-10K.}
For REUTERS-10K, we follow the preprocessing protocol of DEC \citep{xie2016unsupervised}. We use four root categories (corporate/industrial, government/social, markets, and economics) as labels and exclude documents with multiple labels. Following DEC's protocol, we also compute TF--IDF features using the 2,000 most frequent words and use them as input. Due to computational constraints, we randomly sample 10,000 examples. We refer to this sampled dataset as REUTERS-10K. For this dataset, the encoder architecture is
\[
2000 \rightarrow 500 \rightarrow 500 \rightarrow 2000 \rightarrow 10,
\]
with a symmetric decoder and a 10-dimensional latent representation.

\subsection{Implementation Details}
\label{app:implementation_details}

For centroid-based baselines, including k-FED, FKM, and DWF k-Means, we apply the algorithms directly to the raw input features, following the evaluation setting of the original federated clustering papers. Thus, for image datasets, these methods operate on flattened raw pixels, while for REUTERS-10K they operate on TF--IDF features. Deep clustering methods instead learn an autoencoder representation before or during clustering refinement. All federated deep clustering methods use FedAvg as the federated optimization framework. For methods incorporating the sample stability loss or the balanced sampler, each client updates the sampling weights or calculates the Otsu threshold $\tau_k$ once per epoch across its entire local dataset.

\begin{figure}[h]
    \centering
    \includegraphics[width=0.75\textwidth]{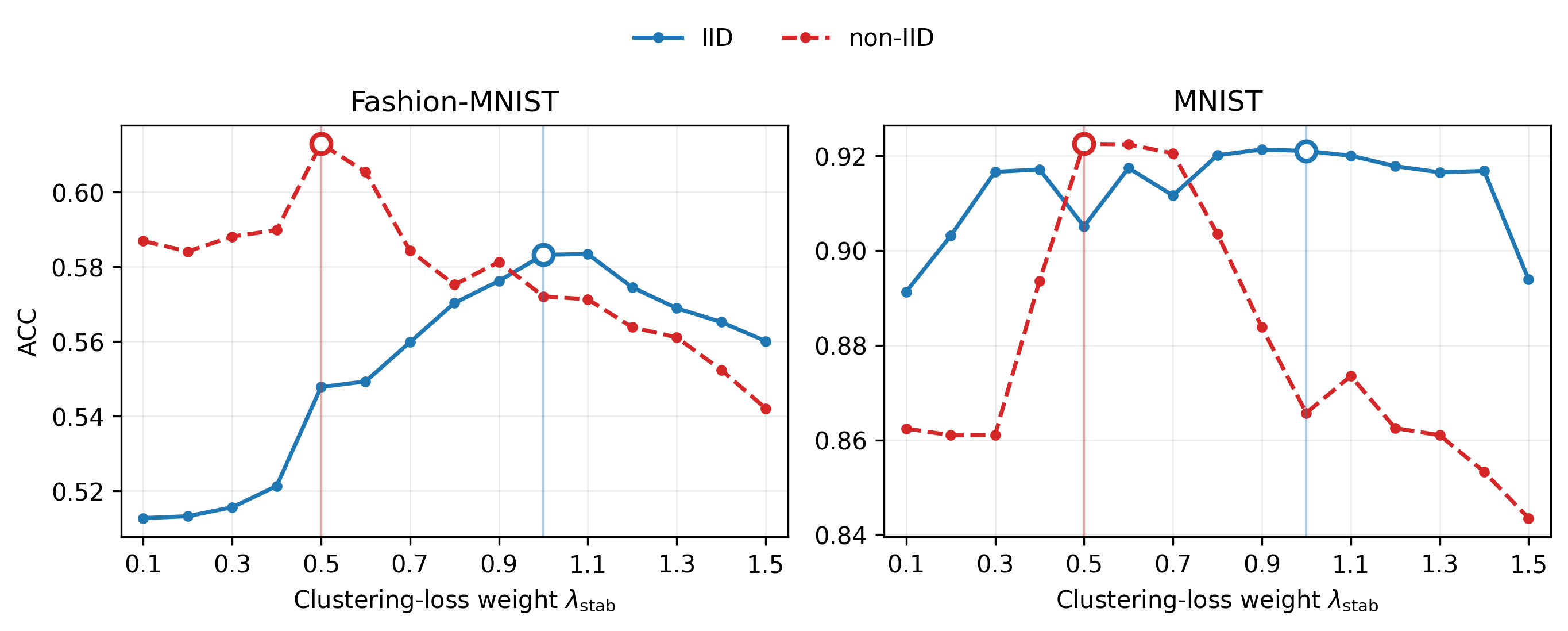}
    \caption{Sensitivity of Fed-BRDECS to the clustering-loss weight $\lambda_{\mathrm{stab}}$ on Fashion-MNIST and MNIST.}
    \label{fig:lambda_clu_sensitivity}
\end{figure}

\paragraph{Selection of loss weights.}
We examine the sensitivity of Fed-BRDECS to the clustering-loss weight $\lambda_{\mathrm{stab}}$ on MNIST and Fashion-MNIST. As shown in Figure~\ref{fig:lambda_clu_sensitivity}, the preferred weight differs between the IID and non-IID settings. We use $\lambda_{\mathrm{stab}}=1.0$ for IID partitions, which provides near-optimal performance on both datasets, and $\lambda_{\mathrm{stab}}=0.5$ for non-IID partitions, which achieves the highest ACC on both datasets. These values are used for the corresponding partition settings in all main experiments. We set $\lambda_{\mathrm{var}}=0.8$, following \citet{cheng2024deep}.

\paragraph{Optimization details.}
Local client optimization uses SGD with momentum. We use a cosine annealing learning-rate scheduler for both the pretraining and refinement stages. The scheduler decays the learning rate from the selected initial value to $10^{-3}$ times the initial value by the final round. For refinement with multiple restarts, the scheduler is applied over the entire refinement process rather than being reset at each restart. We tune weight decay over $\{10^{-4},10^{-5},0\}$ and batch size over $\{64,128,256\}$. The initial learning rate is selected by grid search from $\{1.0,0.2,0.1,0.02,0.01\}$. For local training, each client performs 20 local optimization steps per communication round. All experiments are implemented in PyTorch 2.1.0 with CUDA 12.6. Jobs are executed on four NVIDIA GeForce RTX 2080 Ti GPUs.

\subsection{Full Clustering Results}
\label{app:full_clustering_results}

Table~\ref{tab:main_results} reports the full numerical clustering results corresponding to Figure~\ref{fig:performance_comparison} in the main body. All ACC and NMI values are averaged over five independent runs with different random seeds.

\begin{table}[h]
\centering
\setlength{\tabcolsep}{3.2pt}
\caption{Full clustering results on IID and non-IID federated partitions with $K=10$ clients. Each entry reports ACC and NMI in parentheses. The non-IID setting uses a Dirichlet partition with $\alpha=0.5$. All values are averaged over five independent runs with different random seeds. Higher is better. Best ACC results are shown in bold, and best NMI results are bolded in parentheses. Lower performance of DEC-based refinement compared with Fed-AE indicates that the refinement stage can degrade the pretrained representation, especially under non-IID client distributions.}
\label{tab:main_results}
\vspace{0.2cm}
\resizebox{\textwidth}{!}{%
\begin{tabular}{lcccccccccc}
    \toprule
    Dataset / Partition
    & k-FED
    & FKM
    & DWF
    & CCFC
    & Fed-AE
    & F-DEC
    & Fed-DECS
    & Fed-BDECS
    & Fed-BRDECS
    & Fed-BRIDEC \\
    \midrule
    MNIST / IID
    & 0.4530 (0.4990) & 0.4866 (0.4347) & 0.4914 (0.4318) & 0.8576 (0.8345) & 0.8155 (0.7531) & 0.8878 (\textbf{0.9259}) & 0.8394 (0.7853) & 0.8408 (0.7862) & \textbf{0.9210} (0.8782) & 0.9010 (0.8979) \\
    MNIST / Non-IID
    & 0.5524 (0.4896) & 0.5130 (0.4563) & 0.4972 (0.4338) & 0.8026 (0.7761) & 0.7973 (0.7296) & 0.7016 (0.8062) & 0.8215 (0.7638) & 0.8215 (0.7703) & \textbf{0.9225} (\textbf{0.8498}) & 0.7158 (0.5969) \\
    Fashion-MNIST / IID
    & 0.5297 (0.5228) & 0.4964 (0.5091) & 0.4745 (0.5170) & 0.5448 (0.5798) & 0.5409 (0.5618) & 0.5538 (0.5352) & 0.5438 (0.5748) & 0.5477 (0.5732) & \textbf{0.5832} (\textbf{0.5889}) & 0.5576 (0.5865) \\
    Fashion-MNIST / Non-IID
    & 0.4976 (0.4959) & 0.4987 (0.5322) & 0.5446 (0.5068) & 0.5231 (0.5683) & 0.5614 (0.5797) & 0.5024 (0.5670) & 0.5508 (0.5824) & 0.5685 (0.5764) & \textbf{0.6129} (\textbf{0.5925}) & 0.5438 (0.5805) \\
    KMNIST / IID
    & 0.3066 (0.2790) & 0.3798 (0.2667) & 0.3421 (0.2521) & 0.4685 (0.4345) & 0.4642 (0.4294) & 0.4722 (0.4503) & 0.4735 (0.4267) & 0.4745 (0.4328) & \textbf{0.4798} (0.4381) & 0.4735 (\textbf{0.4405}) \\
    KMNIST / Non-IID
    & 0.3100 (0.2574) & 0.3059 (0.2709) & 0.3352 (0.2656) & 0.3954 (0.3725) & 0.4099 (0.3678) & 0.3854 (0.3932) & 0.3765 (0.3724) & 0.4122 (0.3745) & \textbf{0.4338} (\textbf{0.4065}) & 0.3978 (0.3899) \\
    REUTERS-10K / IID
    & 0.6873 (0.3945) & 0.7590 (0.5098) & 0.7600 (0.4400) & 0.7716 (0.5630) & 0.7285 (0.4904) & 0.7606 (0.5580) & 0.7536 (0.5387) & 0.7559 (0.5742) & \textbf{0.8039} (\textbf{0.5749}) & 0.7832 (0.5439) \\
    REUTERS-10K / Non-IID
    & 0.5962 (0.3316) & 0.7064 (0.4235) & 0.6553 (0.3564) & 0.7528 (0.5274) & 0.7613 (0.5137) & 0.7622 (0.5248) & 0.7658 (0.4984) & 0.7805 (0.4841) & \textbf{0.7865} (\textbf{0.5316}) & 0.7654 (0.4952) \\
    \bottomrule
\end{tabular}%
}
\end{table}

\section{ADDITIONAL DETAILS FOR FEDERATED TIME-SERIES ANOMALY DETECTION}
\label{appx:tsad_details}

\subsection{Related Work}
\label{appx:tsad_related}

Centralized TSAD spans deep one-class methods (Deep SVDD \citep{ruff2018deep}), reconstruction and forecasting methods (LSTM-AE \citep{malhotra2016lstm}, OmniAnomaly \citep{su2019robust}, USAD \citep{audibert2020usad}), graph-based methods (GDN \citep{deng2021graph}), and transformer-based detectors (TranAD \citep{tuli2022tranad}). A growing thread also builds detectors on deep embedded clustering itself, including DTC \citep{madiraju2018deep} and DCLOP \citep{obied2023deep}. Such DEC-based detectors are common in centralized learning but scarce in federated learning, where FedTADBench \citep{liu2022fedtadbench} evaluates representative TSAD algorithms under federated learning. Fed-BRDECS fills this gap as a federated deep clustering backbone for unsupervised outlier detection, whose mechanism can be easily plugged into existing federated TSAD algorithms.

\subsection{Implementation Details}
\label{appx:tsad_impl}

\paragraph{General strategy to instantiate Fed-BRDECS.}
Let $x_{k,i}\in\mathbb{R}^{W\times d}$ denote a time-series window from client $k$, where $W$ is the window length and $d$ is the dimensionality. Let $\mathcal B_k\subset \mathcal D_k$ denote a mini-batch of local windows sampled on client $k$. A reconstruction-based TSAD backbone maps $x_{k,i}$ to a latent representation $h_{k,i}$ and reconstructs the input as $\hat{x}_{k,i}$. Its native training objective is the reconstruction loss
\[
\mathcal{L}_{\mathrm{rec}}^{(k)}
=
\frac{1}{|\mathcal{B}_k|}
\sum_{x_{k,i}\in\mathcal{B}_k}
\ell_{\mathrm{rec}}(x_{k,i},\hat{x}_{k,i}),
\]
where $\ell_{\mathrm{rec}}$ denotes the backbone-specific reconstruction loss. In our experiments, we use the mean squared reconstruction error over the window,
\[
\ell_{\mathrm{rec}}(x_{k,i},\hat{x}_{k,i})
=
\frac{1}{Wd}
\left\|x_{k,i}-\hat{x}_{k,i}\right\|_F^2.
\]
At inference time, the outlier score is computed from the same reconstruction error.

To instantiate Fed-BRDECS, we attach a lightweight projection head $p_\omega$ to the backbone latent representation
\[
z_{k,i}=p_\omega(h_{k,i})\in\mathbb{R}^{d_c},
\]
where $d_c$ is the clustering embedding dimension. Fed-BRDECS then maintains centroids $\{\mu_j\}_{j=1}^{C}$ in this clustering space. During training, soft assignments $q_{kij}$ are computed from $z_{k,i}$ and the centroids using the same Student-$t$ kernel as in the main method. The local objective becomes
\[
\mathcal{L}_{\mathrm{TSAD}}^{(k)}
=
\mathcal{L}_{\mathrm{rec}}^{(k)}
+
\lambda_{\mathrm{stab}}\mathcal{L}_{\mathrm{stab}}^{(k)}.
\]

Centroids are initialized and refreshed by k-FED on the client clustering embeddings. The prediction-balanced sampler is constructed from the induced pseudo-clusters but with smoothing parameterized by $\gamma$. Thus, windows assigned to locally rare predicted regimes are sampled more often, while the smoothing parameter $\gamma$ controls the strength of rebalancing and prevents the sampler from over-concentrating on a small number of windows.

At inference time, the projection head and centroids are discarded, and outlier scores are computed only from the original reconstruction error. Therefore, Fed-BRDECS acts as a training-time representation-learning module and does not add inference-time cost.

\paragraph{LSTM-AE instantiation.}
For LSTM-AE, the encoder maps each input window $x_{k,i}$ to a final recurrent hidden state $h_{k,i}$. We attach the clustering projection head to this final hidden state:
\[
z_{k,i}=p_\omega(h_{k,i}).
\]
The decoder reconstructs the input window, and the reconstruction loss is the mean squared error between $x_{k,i}$ and $\hat{x}_{k,i}$. Thus, the LSTM-AE backbone is unchanged except for the temporary clustering head used during training. Since $h_{k,i}$ is already a single vector representation of the full window, no temporal pooling is required before clustering.

\paragraph{TranAD instantiation.}
TranAD uses a two-phase transformer reconstruction mechanism. Given an input window, the first decoder produces an initial reconstruction, and the second decoder reconstructs the target using the first-phase reconstruction error as additional context. We attach the clustering head to the latent representation produced by the first-phase transformer encoder. Let
\[
M_{k,i}^{(1)}\in\mathbb{R}^{W\times d_h}
\]
denote the first-phase encoder memory for window $x_{k,i}$. We obtain a window-level representation by mean pooling over time:
\[
h_{k,i}
=
\frac{1}{W}
\sum_{t=1}^{W}
M_{k,i,t}^{(1)}.
\]
The clustering embedding is then
\[
z_{k,i}=p_\omega(h_{k,i}).
\]
The native TranAD reconstruction loss is kept unchanged. In particular, if $x_{k,i}^{(1)}$ and $x_{k,i}^{(2)}$ denote the first- and second-phase reconstructions, the local reconstruction loss uses the same epoch-dependent weighting as TranAD:
\[
\ell_{\mathrm{rec}}
=
\frac{1}{e}
\ell(x_{k,i}^{(1)},x_{k,i})
+
\left(1-\frac{1}{e}\right)
\ell(x_{k,i}^{(2)},x_{k,i}),
\]
where $e$ is the local epoch index. At inference time, the outlier score is computed from the second-phase reconstruction error, as in the original TranAD.

\paragraph{Prediction-balanced sampling for TSAD.}
For each client, Fed-BRDECS periodically constructs a prediction-balanced sampler from the pseudo-clusters induced by the temporary clustering head. The purpose is to increase the exposure of locally rare normal regimes during training, while avoiding overly aggressive oversampling that may overfit a small number of anomalous windows.

Specifically, client $k$ computes soft assignments over its local windows and estimates the local soft cluster mass
\[
f_{k,j}
=
\sum_{x_{k,i}\in \mathcal D_k}
q_{kij},
\qquad j=1,\ldots,C.
\]
Each window is assigned to its most likely pseudo-cluster,
\[
\hat y_{k,i}
=
\arg\max_{j\in\{1,\ldots,C\}}
q_{kij}.
\]
The inverse-frequency cluster weight is then smoothed by an exponent $\gamma$:
\[
\tilde w_{k,j}
=
\left(
\frac{1}{f_{k,j}+\epsilon}
\right)^\gamma,
\]
and each window receives the sampling weight
\[
w_{k,i}
=
\tilde w_{k,\hat y_{k,i}}.
\]
Mini-batches are sampled with replacement according to probabilities proportional to $w_{k,i}$. Here, $\gamma$ controls the strength of rebalancing: $\gamma=0$ recovers uniform sampling, while $\gamma=1$ corresponds to the standard inverse-frequency sampler. This smoothing is used to prevent the sampler from concentrating too heavily on rare pseudo-clusters, which may otherwise increase the risk of overfitting to anomalous or atypical windows. In our TSAD experiments, we use $\gamma=0.5$.

\paragraph{Centroid initialization and restarting.}
After reconstruction pretraining, each client extracts clustering embeddings $\{z_{k,i}\}$ from its local windows and runs local k-means. The server aggregates the returned local centroids using k-FED to obtain the initial global centroids. During refinement, Fed-BRDECS periodically refreshes centroids by rerunning the same k-FED procedure on the current clustering embeddings. This centroid-level restarting corrects biased or inactive centroids caused by heterogeneous local time-series regimes. No clustering head is used at inference.

\paragraph{Details for proprietary sensor dataset.}
The proprietary sensor dataset is collected from industrial manufacturing equipment and consists of irregularly sampled multivariate time-series samples. After preprocessing, the dataset contains $72{,}808$ samples with an anomaly rate of $3.12\%$. The anomaly labels are used only for evaluation and are not used during unsupervised federated training.

\begin{figure}[h]
    \centering
    \includegraphics[width=0.55\textwidth]{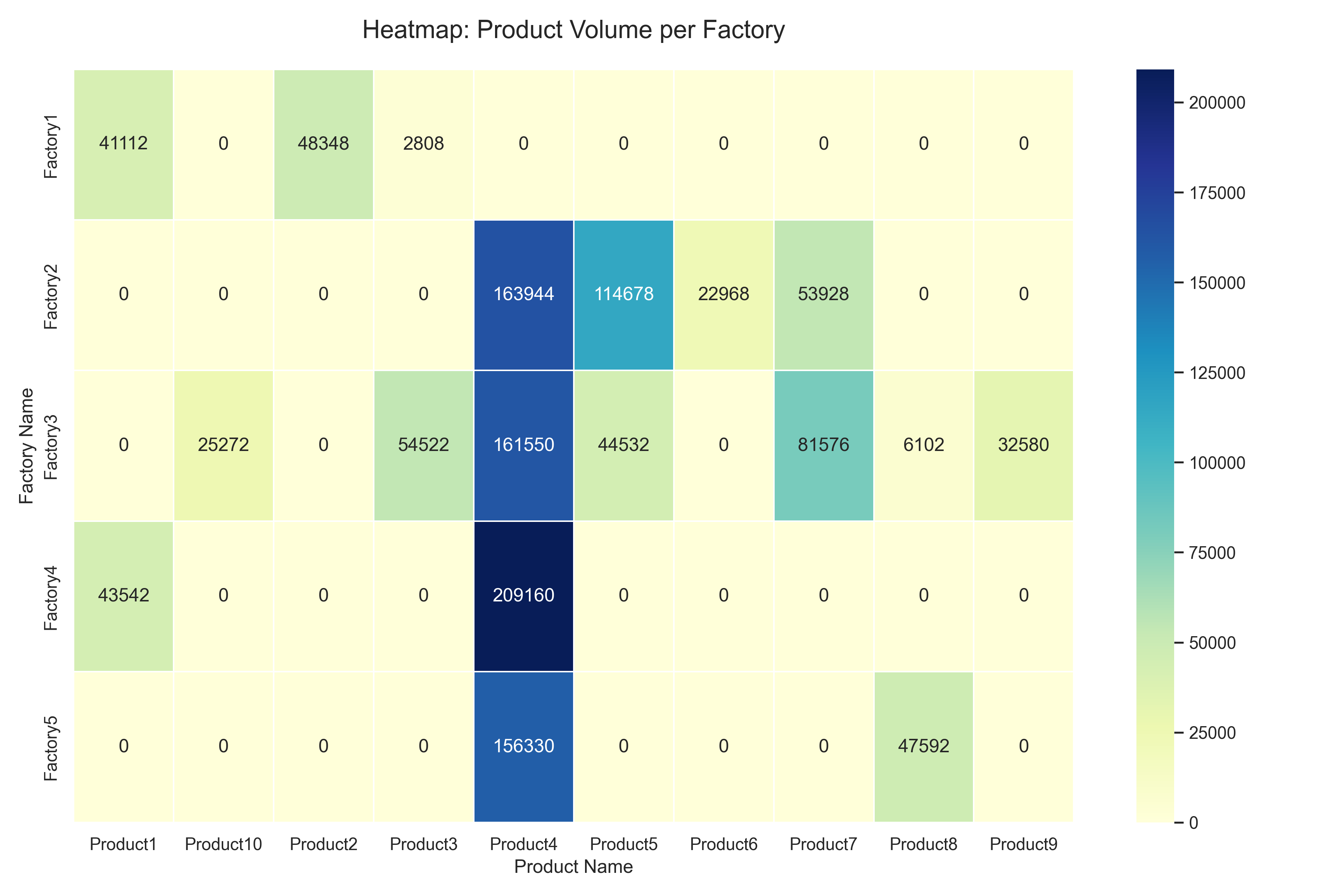}
    \caption{Factory--product production-volume metadata for the proprietary sensor dataset. The heatmap shows strong non-IID structure across factories.}
    \label{fig:sensor_factory_product_heatmap}
\end{figure}

The dataset is naturally federated by factory, where each factory is treated as one client. Figure~\ref{fig:sensor_factory_product_heatmap} summarizes the factory--product metadata. The metadata covers five factories and ten products, but the support is sparse, indicating the strong client heterogeneity.

\begin{figure}[h]
    \centering
    \includegraphics[width=0.8\textwidth]{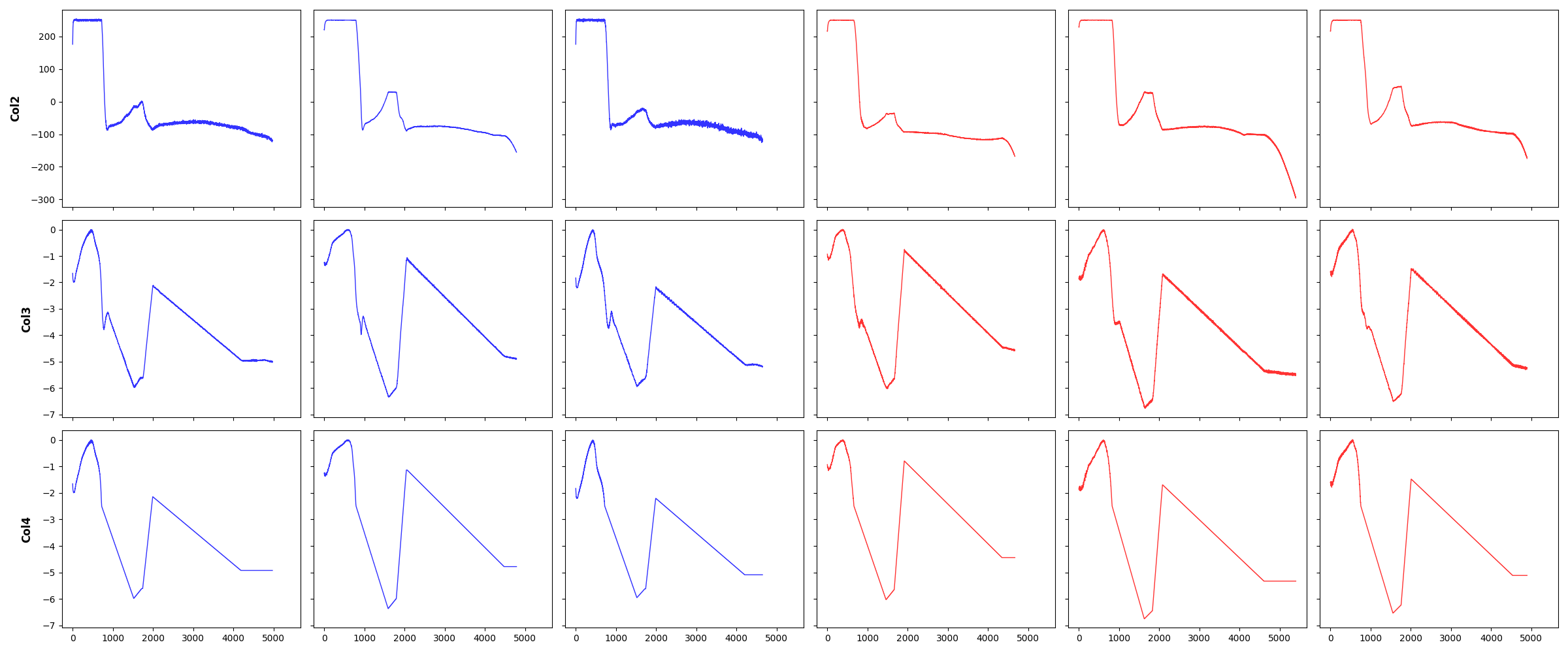}

    \vspace{2mm}

    \includegraphics[width=0.8\textwidth]{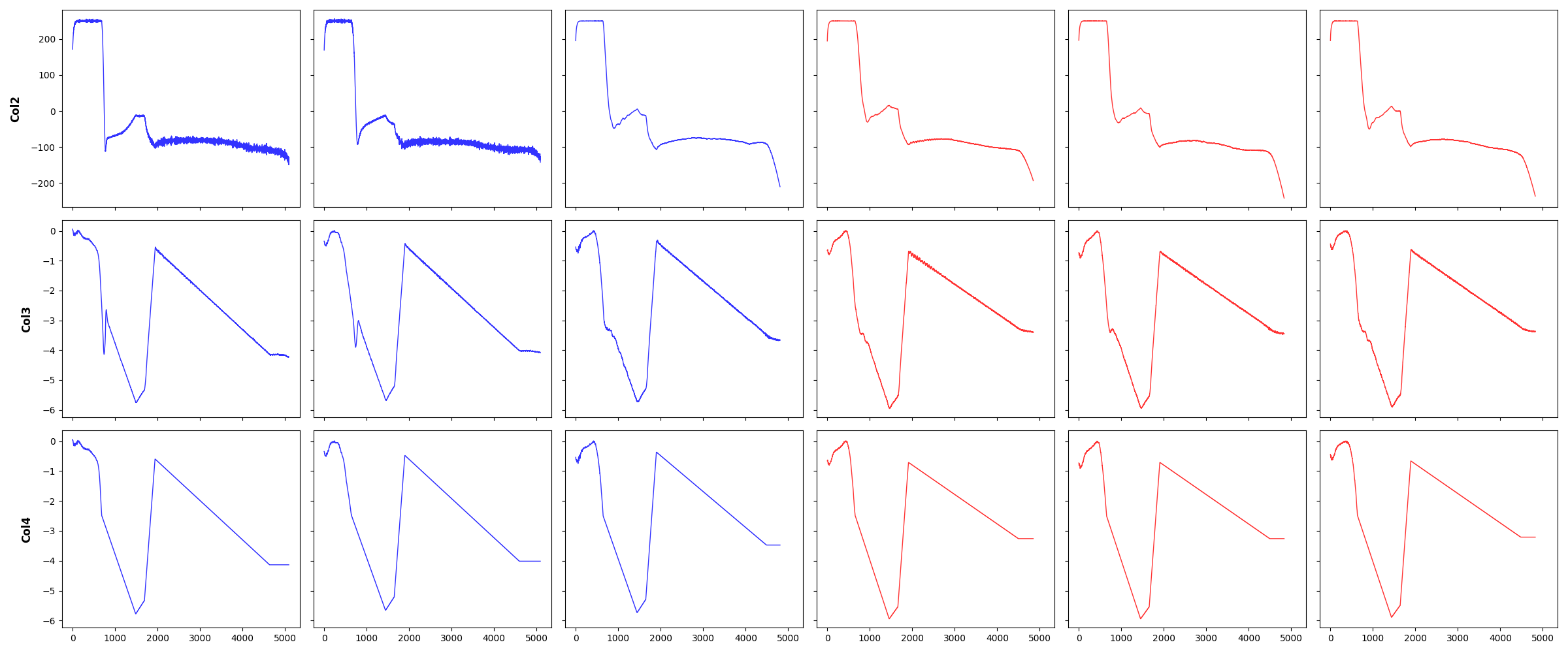}

    \vspace{2mm}

    \includegraphics[width=0.8\textwidth]{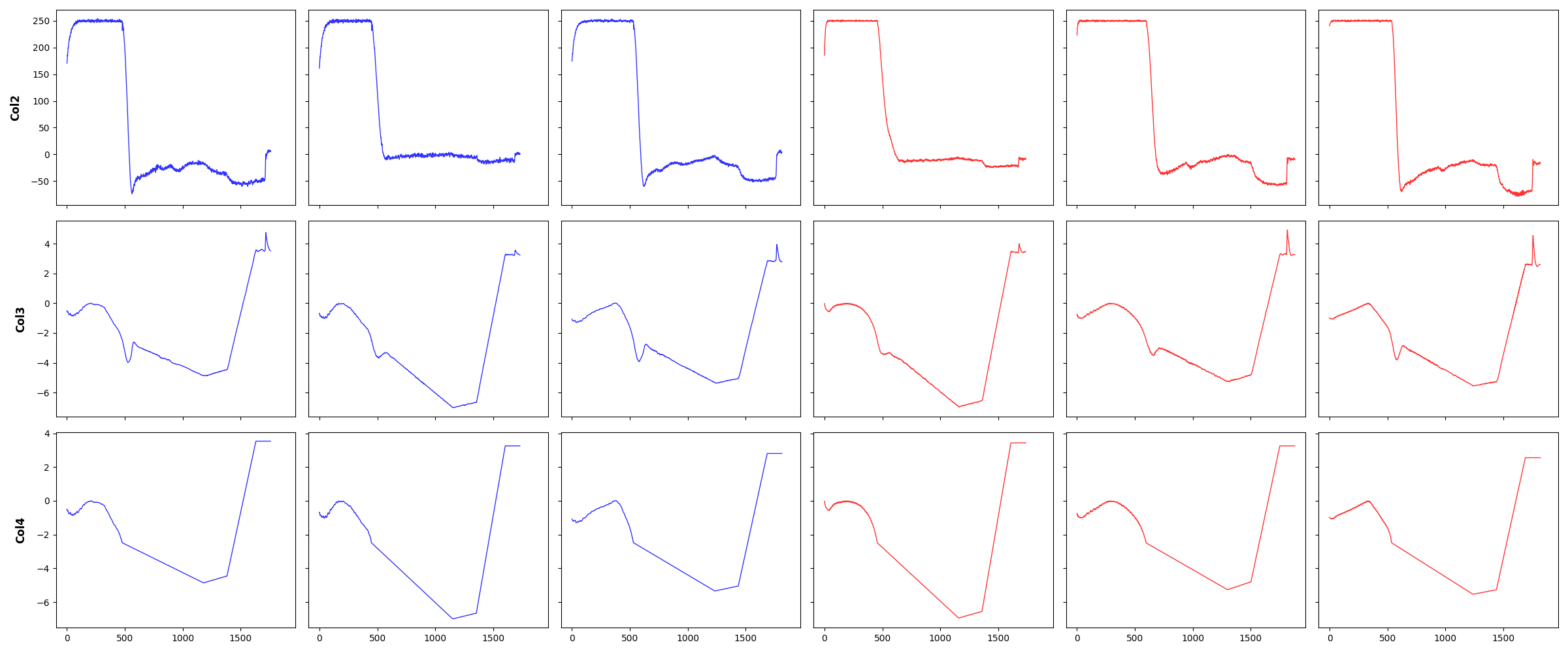}
    \caption{Representative multivariate sensor trajectories from Products~1, 2, and~4 in the proprietary dataset. Each panel visualizes multiple examples over three sensor channels. The examples show product-dependent temporal structure, variable sequence length, sharp transitions, plateau regions, and heterogeneous channel dynamics, motivating the use of CoFormer-AE for irregularly sampled sensor data.}
    \label{fig:sensor_product_examples}
\end{figure}

Figure~\ref{fig:sensor_product_examples} shows representative multivariate trajectories from three product types. The trajectories exhibit irregular temporal structure, product-dependent sequence lengths, sharp transitions, and different sensor-channel dynamics. These observations support the use of a model designed for irregularly sampled time series.

\paragraph{CoFormer-AE for irregularly sampled sensor data.}
The proprietary sensor dataset consists of irregularly sampled multivariate time series. Therefore, we use CoFormer-AE instead of LSTM-AE or TranAD for this dataset. CoFormer provides the window-level representation $h_{k,i}$ used by the reconstruction model, and Fed-BRDECS is attached in the same way,
\[
z_{k,i}=p_\omega(h_{k,i}).
\]
The remaining procedure, including k-FED centroid initialization, sample-stability loss, prediction-balanced sampling, and inference by reconstruction error, is unchanged. We do not evaluate TranAD on this dataset because TranAD is architecture-specific to regularly sampled fixed-window inputs and is not directly compatible with the CoFormer backbone.

\end{document}